\documentclass[10pt,twocolumn,letterpaper]{article}
\pdfoutput=1
\usepackage{cvpr}
\usepackage{times}
\usepackage{epsfig}
\usepackage{graphicx}
\usepackage{epstopdf}
\usepackage{amsmath}
\usepackage{amssymb}
\usepackage{amsthm}
\usepackage{enumerate}
\usepackage{subcaption}
\usepackage{multirow}
\usepackage{pbox}
\usepackage{array}
\usepackage{comment}

\usepackage{rotating}
\usepackage{capt-of}% or \usepackage{caption}

\usepackage[pagebackref=true,breaklinks=true,letterpaper=true,colorlinks,bookmarks=false,
			pdfauthor={D. Khue Le-Huu},
			pdftitle={Alternating Direction Graph Matching},
			pdfsubject={Computer Vision and Pattern Recognition},
			pdfkeywords={graph matching, hypergraph matching, higher-order graphs, admm, nonconvex, decomposition}]{hyperref}

\cvprfinalcopy % *** Uncomment this line for the final submission

\newtheoremstyle{note}% name
{3pt}% Space above
{3pt}% Space below
{\itshape}% Body font
{}% Indent amount
{\bfseries}% Theorem head font
{}% Punctuation after theorem head
{.5em}% Space after theorem head
{}% Theorem head spec (can be left empty, meaning ‘normal’)
\theoremstyle{note}

\newtheorem*{lemma*}{Lemma}

\newtheorem*{subproblem*}{Subproblem}
\newtheorem{problem}{Problem}
\newtheorem*{problem*}{Problem}

\newtheorem*{conjecture*}{Conjecture}

\newtheorem*{definition*}{Definition}

\newcommand{\argmin}{\operatornamewithlimits{argmin}}
\newcommand{\set}[1]{\left\{ #1 \right\}}
\newcommand{\abs}[1]{\left| #1 \right|}
\newcommand{\norm}[1]{\left\| #1 \right\|_2}

\newcommand{\cE}{\mathcal{E}}
\newcommand{\cF}{\mathcal{F}}
\newcommand{\cG}{\mathcal{G}}

\newcommand{\cM}{\mathcal{M}}

\newcommand{\cV}{\mathcal{V}}
\newcommand{\cX}{\mathcal{X}}

\newcommand{\RR}{\mathbb{R}}

\renewcommand{\c}{\mathbf{c}}

\newcommand{\p}{\mathbf{p}}
\newcommand{\s}{\mathbf{s}}

\renewcommand{\v}{\mathbf{v}}
\newcommand{\x}{\mathbf{x}}
\newcommand{\y}{\mathbf{y}}
\newcommand{\z}{\mathbf{z}}

\newcommand{\A}{\mathbf{A}}

\newcommand{\M}{\mathbf{M}}
\newcommand{\V}{\mathbf{V}}
\newcommand{\X}{\mathbf{X}}

\newcommand{\Mc}{\mathcal{M}_\mathrm{c}}

\newcommand{\Mr}{\mathcal{M}_\mathrm{r}}

\newcommand{\mat}{\mathrm{mat}}
\newcommand{\vect}{\mathrm{vec}}

\def\cvprPaperID{2763} % *** Enter the CVPR Paper ID here

\ifcvprfinal\fi
\begin{document}
%\setlength{\belowdisplayskip}{5pt} 
%\setlength{\belowdisplayshortskip}{3pt}
%\setlength{\abovedisplayskip}{5pt} 
%\setlength{\abovedisplayshortskip}{3pt}
%%%%%%%%% TITLE
\title{Alternating Direction Graph Matching}

\author{D. Khu\^e L\^e-Huu\textsuperscript{1, 2} \qquad Nikos Paragios\textsuperscript{1, 2, 3}\\
\normalsize\textsuperscript{1}CentraleSup\'elec, Universit\'e Paris-Saclay\qquad \textsuperscript{2}Inria\qquad\textsuperscript{3}TheraPanacea
\\
{\tt\small \{khue.le, nikos.paragios\}@centralesupelec.fr}
}

\maketitle
%\thispagestyle{empty}

%%%%%%%%% ABSTRACT
\begin{abstract}\vspace{-4pt}
   In this paper, we introduce a graph matching method that can account for constraints of arbitrary order, with arbitrary potential functions. Unlike previous decomposition approaches that rely on the graph structures, we introduce a decomposition of the matching constraints. Graph matching is then reformulated as a non-convex non-separable optimization problem that can be split into smaller and much-easier-to-solve subproblems, by means of the alternating direction method of multipliers. The proposed framework is modular, scalable, and can be instantiated into different variants. Two instantiations are studied exploring pairwise and higher-order constraints. Experimental results on widely adopted benchmarks involving synthetic and real examples demonstrate that the proposed solutions outperform existing pairwise graph matching methods, and competitive with the state of the art in higher-order settings.
\end{abstract}\vspace{-5pt}

%%%%%%%%% BODY TEXT
\section{Introduction}
The task of finding correspondences between two sets of visual features has a wide range of applications in computer vision and pattern recognition. This problem can be effectively solved using graph matching~\cite{torresani2013dual}, and as a consequence, these methods have been successfully applied to various vision tasks, such as stereo matching~\cite{kolmogorov2001computing}, object recognition and categorization~\cite{belongie2002shape, duchenne2011graph}, shape matching~\cite{belongie2002shape}, surface registration~\cite{zeng2010dense}, \etc 

The general idea of solving feature correspondences via graph matching is to associate each set of features an attributed graph, where node attributes describe local characteristics, while edge (or hyper-edge) attributes describe structural relationships. The matching task seeks to minimize an energy (objective) function composed of unary, pairwise, and potentially higher-order terms. These terms are called the \emph{potentials} of the energy function. In pairwise settings, graph matching can be seen as a quadratic assignment problem (QAP) in general form, known as Lawler's QAP~\cite{lawler1963quadratic}. Since QAP is known to be NP-complete~\cite{burkard1998quadratic,sahni1976p}, graph matching is also NP-complete~\cite{gold1996graduated} and only approximate solutions can be found in polynomial time.

Graph matching has been an active research topic in the computer vision field for the past decades. In the recent literature,~\cite{gold1996graduated} proposed a graduated assignment algorithm to iteratively solve a series of convex approximations to the matching problem. In~\cite{leordeanu2005spectral}, a spectral matching based on the rank-$1$ approximation of the \emph{affinity matrix} (composed of the potentials) was introduced, which was later improved in~\cite{cour2007balanced} by incorporating affine constraints towards a tighter relaxation. In~\cite{leordeanu2009integer}, an integer projected fixed point algorithm that solves a sequence of first-order Taylor approximations using Hungarian method~\cite{kuhn1955hungarian} was proposed, while in~\cite{torresani2013dual} the dual of the matching problem was considered to obtain a lower-bound on the energy, via dual decomposition. In~\cite{cho2010reweighted}, a random walk variant was used to address graph matching while~\cite{zhou2012factorized} factorized the affinity matrix into smaller matrices, allowing a convex-concave relaxation that can be solved in a path-following fashion. Their inspiration was the path-following approach~\cite{zaslavskiy2009path} exploiting a more restricted graph matching formulation, known as Koopmans-Beckmann's QAP~\cite{koopmans1957assignment}. Lately,~\cite{cho2014finding} proposed a max-pooling strategy within the graph matching framework that is very robust to outliers.

Recently, researchers have proposed higher-order graph matching models to better incorporate structural similarities and achieve more accurate results~\cite{duchenne2011tensor,zass2008probabilistic}. For solving such high-order models,~\cite{zass2008probabilistic} viewed the matching problem as a probabilistic model that is solved using an iterative successive projection algorithm. The extension of pairwise methods to deal with higher-order potentials was also considered like for example in~\cite{duchenne2011tensor} through a tensor matching (extended from~\cite{leordeanu2005spectral}), or in~\cite{zeng2010dense} through a third-order dual decomposition (originating from~\cite{torresani2013dual}), or in~\cite{lee2011hyper} through a high-order reweighted random walk matching (extension of~\cite{cho2010reweighted}). Recently,~\cite{nguyen2015flexible} developed a block coordinate ascent algorithm for solving third-order graph matching. They lifted the third-order problem to a fourth-order one which, after a convexification step, is solved by a sequence of linear or quadratic assignment problems. Despite the impressive performance, this method has two limitations: (a) it cannot be applied to graph matching of arbitrary order other than third and fourth, and (b) it cannot deal with graph matching where occlusion is allowed on both sides, nor with many-to-many matching.

In this paper, a novel class of algorithms is introduced for solving graph matching involving constraints with arbitrary order and arbitrary potentials. These algorithms rely on a decomposition framework using the alternating direction method of multipliers.

The remainder of this paper is organized as follows. Section~\ref{sec:background} provides the mathematical foundations of our approach while in Section~\ref{sec:adgm} the general decomposition strategy is proposed along with two instantiations of this framework to a pairwise and higher-order approach. Section~\ref{sec:experiments} presents in-depth experimental validation and comparisons with competing methods. The last section concludes the paper and presents the perspectives.

\section{Mathematical background and notation}\label{sec:background}
Let us first provide the elementary notation as well as the basic mathematical foundations of our approach. In the first subsection we will give a brief review of tensor, which will help us to compactly formulate the graph matching problem, as will be shown in the subsequent subsection.

\subsection{Tensor}
A real-valued $D\textsuperscript{th}$-order tensor $\cF$ is a multidimensional array belonging to $\RR^{n_1\times n_2 \times\cdots \times n_D}$ (where $n_1,n_2,\ldots,n_D$ are positive integers). We denote the elements of $\cF$ by $\cF_{i_1i_2\ldots i_D}$ where $1\le i_d \le n_d$ for $d=1,2,\ldots, D$. Each dimension of a tensor is called a \emph{mode}. 

We call the \emph{multilinear form} associated to a tensor $\cF$ the function $F:\RR^{n_1}\times \RR^{n_2} \times\cdots \times \RR^{n_D}\to \RR$ defined by
\begin{equation}\label{eq:multilinear}
F(\x_1,\ldots,\x_D) = \sum_{i_1=1}^{n_1}\cdots\sum_{i_D=1}^{n_D} \cF_{i_1i_2\ldots i_D}x^1_{i_1}x^2_{i_2}\cdots x^D_{i_D}
\end{equation} 
where $\x_d=(x^d_1,x^d_2,\ldots,x^d_{n_d})\in\mathbb{R}^{n_d}$ for $d=1,2,\ldots,D$.

A tensor can be multiplied by a vector at a specific mode. Let $\mathbf{v}=(v_1,v_2,\ldots,v_{n_d})$ be an $n_d$ dimensional vector. The \emph{mode-$d$ product} of $\cF$ and $\v$, denoted by $\cF\otimes_d\v$, is a $(D-1)\textsuperscript{th}$-order tensor $\cG$ of dimensions $n_1\times\cdots  \times n_{d-1}\times n_{d+1}\times\cdots \times n_D$ defined by 
\begin{equation}
\cG_{i_1\ldots i_{d-1}i_{d+1}\ldots i_D} = \sum_{i_d=1}^{n_d} \cF_{i_1\ldots i_{d}\ldots i_D}v_{i_d}.
\end{equation}
With this definition, it is straightforward to see that the multilinear form~\eqref{eq:multilinear} can be re-written as
\begin{equation}\label{eq:multilinear-tensor}
F(\x_1,\x_2,\ldots,\x_D)  = \cF\otimes_1 \x_1 \otimes_2 \x_2\cdots\otimes_D\x_D.
\end{equation}
Let us consider for convenience the notation $\bigotimes_{d=a}^b$ to denote a sequence of products from mode $a$ to mode $b$:
\begin{equation}\label{eq:bigotimes}
\cF\bigotimes_{d=a}^b\x_d  = \cF\otimes_a \x_a \otimes_{a+1} \x_{a+1}\cdots\otimes_b\x_b.
\end{equation}
By convention, $\cF\bigotimes_{d=a}^b\x_d = \cF$ if $a > b$.

In this work, we are interested in tensors having the same dimension at every mode, \ie $n_1=n_2=\ldots = n_D =n$. In the sequel, all tensors are supposed to have this property.

\subsection{Graph and hypergraph matching}\label{sec:graph-and-hypergraph-matching}
A matching configuration between two graphs $\cG_1=(\cV_1,\cE_1)$ and $\cG_2=(\cV_2,\cE_2)$ can be represented by a \emph{assignment matrix} $\X \in\set{0,1}^{n_1\times n_2}$ where $n_1=\abs{\cV_1}, n_2=\abs{\cV_2}$. An element $x_{(i_1,i_2)}$ of $\X$ equals $1$ if the node $i_1\in\cV_1$ is matched to the node $i_2\in\cV_2$, and equals $0$ otherwise.\\ 
Standard graph matching imposes the one-to-(at most)-one constraints, \ie the sum of any row or any column of $\X$ must be $\le 1$. If the elements of $\X$ are binary, then $\X$ obeys the \emph{hard matching constraints}. When $\X$ is relaxed to take real values in $[0,1]$, $\X$ obeys the \emph{soft matching constraints}.

In this paper we use the following notations: $\vect(\V)$ denotes the column-wise vectorized replica of a matrix $\V$; $\mat(\v)$ is the $n_1\times n_2$ reshaped matrix of an $n$-dimensional vector $\v$, where $n = n_1n_2$; $\X \in\RR^{n_1\times n_2}$ the assignment matrix and $\x= \vect(\X) \in\RR^{n}$ the \emph{assignment vector}; $\cM^*$ (respectively $\cM$) is the set of $n_1\times n_2$ matrices that obey the hard (respectively, the soft) matching constraints.\\

\noindent\textbf{Energy function.} Let $x_i = x_{(i_1,i_2)}$ be an element of $\X$ representing the matching of two nodes $i_1$ and $i_2$. Suppose that matching these nodes requires a potential $\cF_i^1 \in\RR$. Similarly, let $\cF^2_{ij}$ denote the potential for matching two edges $(i_1,j_1)$, $(i_2,j_2)$, and $\cF^3_{ijk}$ for matching two (third-order) hyper-edges $(i_1,j_1,k_1)$, $(i_2,j_2,k_2)$, and so on. Graph matching can be expressed as minimizing
\begin{equation}\label{eq:energy-1}
\sum_{i} \cF^1_i x_i + \sum_{ij} \cF^2_{ij}x_ix_j + \sum_{ijk} \cF^3_{ijk}x_ix_jx_k + \cdots
\end{equation}
The above function can be re-written more compactly using tensors. Indeed, let us consider for example the third-order potentials. Since $\cF^3_{ijk}$ has three indices, $(\cF^3_{ijk})_{1\le i,j,k\le n}$ can be seen as a third-order tensor belonging to $\RR^{n\times n\times n}$ and its multilinear form (\cf \eqref{eq:multilinear}) is the function
\begin{equation}
F^3(\x,\y,\z) = \sum_{i=1}^n\sum_{j=1}^n\sum_{k=1}^n \cF^3_{ijk}x_iy_jz_k
\end{equation}
defined for $\x,\y,\z\in\RR^n$. Clearly, the third-order terms in~\eqref{eq:energy-1} can be re-written as $F^3(\x,\x,\x)$. More generally, $D\textsuperscript{th}$-order potentials can be represented by a $D\textsuperscript{th}$-order tensor $\cF^D$ and their corresponding terms in the objective function can be re-written as $F^D(\x,\x,\ldots,\x)$, resulting in the following reformulation of graph matching.

\begin{problem}[$D\textsuperscript{th}$-order graph matching] \label{prob:generalized-GM}  Minimize
\begin{equation}\label{eq:energy-unhomogeneous}F^1(\x) + F^2(\x,\x) + \cdots + F^D(\x,\x,\ldots,\x)
\end{equation}
subject to $\x\in\cM^*$, where $F^d$ ($d=1,\ldots,D$) is the multilinear form of a tensor $\cF^d$ representing the $d\textsuperscript{th}$-order potentials.
\end{problem}
In the next section, we propose a method to solve the continuous relaxation of this problem, \ie minimizing $\eqref{eq:energy-unhomogeneous}$ subject to $\x\in\cM$ (soft matching) instead of $\x\in\cM^*$ (hard matching). The returned continuous solution is discretized using the usual Hungarian method~\cite{kuhn1955hungarian}.

\section{Alternating direction graph matching}\label{sec:adgm}
\subsection{Overview of ADMM}\label{sec:admm}
We briefly describe the (multi-block) alternating direction method of multipliers (ADMM) for solving the following optimization problem:
\begin{equation}\label{prob:admm}
\begin{aligned}
\mbox{Minimize}\quad &\phi(\x_1,\x_2,\ldots,\x_p)\\ 
\mbox{subject to}\quad & \A_1\x_1 + \A_2\x_2+\cdots + \A_p\x_p = \mathbf{b},\\
& \x_i\in\cX_i \subseteq\RR^{n_i} \quad\forall 1\le i\le p,
\end{aligned}
\end{equation}
where $\cX_i$ are closed convex sets, $\mathbf{A}_i\in\RR^{m\times n_i} \ \forall i,\mathbf{b}\in\RR^{m}$. 

The augmented Lagrangian of the above problem is
\begin{multline}
L_\rho(\x_1,\x_2,\ldots,\x_p,\y) = \phi(\x_1,\x_2,\ldots,\x_p) \\ + \y^\top\left(\sum_{i=1}^p\A_i\x_i- \mathbf{b}\right) + \frac{\rho}{2}\norm{\sum_{i=1}^p\A_i\x_i- \mathbf{b}}^2,
\end{multline}
where $\y$ is called the \emph{Lagrangian multiplier vector} and $\rho > 0$ is called the \emph{penalty parameter}.

In the sequel, let $\x_{[a,b]}$ denote $(\x_a,\x_{a+1},\ldots,\x_b)$ (by convention, if $a>b$ then $\x_{[a,b]}$ is ignored). Standard ADMM solves problem~\eqref{prob:admm} by iterating:
\begin{enumerate}
\item For $i=1,2,\ldots,p$, update $\x_i$:
\begin{equation}\label{eq:x-update-general}
\x_i^{k+1} =\argmin_{\x\in\cX_i} L_\rho(\x_{[1,i-1]}^{k+1},\x,\x_{[i+1,p]}^{k}, \y^k).
\end{equation}
\item Update $\y$:
\begin{equation}\label{eq:y-update-general}
\y^{k+1} =\y^{k} + \rho\left(\sum_{i=1}^p\A_i\x_i^{k+1} - \mathbf{b}^{k+1}\right).
\end{equation}
\end{enumerate}
The algorithm converges if the following \emph{residual} converges to $0$ as $k\to\infty$:
\begin{equation}\label{eq:residual}
r^k = \norm{\sum_{i=1}^{p} \A_i\x_i^{k} - \mathbf{b}^{k}}^2 + \sum_{i=1}^{p}\norm{\A_i\x_i^k - \A_i\x_i^{k-1}}^2.
\end{equation}
We will discuss the convergence of ADMM in Section~\ref{sec:convergent-adgm}.

\subsection{Graph matching decomposition framework}\label{sec:framework}
Decomposition is a general approach to solving a problem by breaking it up into smaller ones that can be efficiently addressed separately, and then reassembling the results towards a globally consistent solution of the original non-decomposed problem~\cite{bertsekas1999nonlinear,boyd2007notes,dantzig1960decomposition}. Clearly, the above ADMM is such a method because it decomposes the large problem~\eqref{prob:admm} into smaller problems~\eqref{eq:x-update-general}.

In computer vision, decomposition methods such as Dual Decomposition (DD) and ADMM have been applied to optimizing discrete Markov random fields (MRFs)~\cite{komodakis2009beyond,komodakis2011mrf,lehuu2014dual,martins2015ad} and to solving graph matching~\cite{torresani2013dual}. The main idea is to decompose the original complex graph into simpler subgraphs and then reassembling the solutions on these subgraphs using different mechanisms. While in MRF inference, this concept has been proven to be flexible and powerful, that is far from being the case in graph matching, due to the hardness of the matching constraints. Indeed, to deal with these constraints,~\cite{torresani2013dual} for example adopted a strategy that creates subproblems that are also smaller graph matching problems, which are computationally highly challenging. Moreover, subgradient method has been used to impose consensus, which is known to have slow rate of convergence~\cite{bertsekas1999nonlinear}. One can conclude that DD is a very slow method and works for a limited set of energy models often associated with small sizes and low to medium geometric connectivities~\cite{torresani2013dual}.

In our framework, we do not rely on the structure of the graphs but instead, on the nature of the variables. In fact, the idea is to decompose the assignment vector $\x$ (by means of Lagrangian relaxation) into different variables where each variable obeys weaker constraints (that are easier to handle). For example, instead of dealing with the assignment vector $\x\in\cM$, we can represent it by two vectors $\x_1$ and $\x_2$, where the sum of each row of $\mat(\x_1)$ is $\le 1$ and the sum of each column of $\mat(\x_2)$ is $\le 1$, and we constrain these two vectors to be equal. More generally, we can decompose $\x$ into as many vectors as we want, and in any manner, the only condition is that the set of constraints imposed on these vectors must be equivalent to $\x_1=\x_2=\cdots=\x_p \in\cM$ where $p$ is the number of vectors. As for the objective function~\eqref{eq:energy-unhomogeneous}, there is also an infinite number of ways to re-write it under the new variables $\x_1,\x_2,\ldots,\x_p$. The only condition is that the re-written objective function must be equal to the original one when $\x_1=\x_2=\cdots=\x_p=\x$. For example, if $p=D$ then one can re-write~\eqref{eq:energy-unhomogeneous} as
\begin{equation}
F^1(\x_1) + F^2(\x_1,\x_2) + \cdots + F^D(\x_1,\x_2,\ldots,\x_D).
\end{equation}
Each combination of (a) such a variable decomposition and (b) such a way of re-writing the objective function will yield a different Lagrangian relaxation and thus, produce a different algorithm. Since there are virtually infinite of such combinations, the number of algorithms one can design from them is also unlimited, not to mention the different choices of the reassembly mechanism, such as subgradient methods~\cite{bertsekas1999nonlinear, boyd2007notes}, cutting plane methods~\cite{bertsekas1999nonlinear}, ADMM~\cite{boyd2011distributed}, or others. We call the class of algorithms that base on ADMM Alternating Direction Graph Matching (ADGM) algorithms. A major advantage of ADMM over the other mechanisms is that its subproblems involve only one block of variables, regardless of the form the objective function.

As an illustration of ADGM, we present below a particular example. Nevertheless, this example is still general enough to include an infinite number of special cases.

\begin{problem}[Decomposed graph matching]\label{prob:GM-decomposed}
Minimize 
\begin{equation}\label{eq:GM-decomposed}
F^1(\x_1) + F^2(\x_1,\x_2) + \cdots + F^D(\x_1,\x_2,\ldots,\x_D)
\end{equation}
subject to
\begin{align}
&\A_1\x_1+\A_2\x_2+\cdots+\A_D\x_D = \mathbf{0}, \label{eq:v0-equal}\\
&\x_d\in\cM_d \quad \forall\ 1\le d \le D, \label{eq:v0-set}
\end{align}
where $(\A_d)_{1\le d\le D}$ are $m\times n$ matrices, defined in such a way that~\eqref{eq:v0-equal} is equivalent to $\x_1=\x_2=\cdots=\x_D$, and $(\cM_d)_{1\le d\le D}$ are closed convex subsets of $\RR^n$ satisfying
\begin{equation}\label{eq:v0-set-condition}
\cM_1\cap\cM_2\cap\cdots\cap\cM_D = \cM. %\bigcap_{d=1}^D\cM_d = \cM.
\end{equation}
\end{problem}
It is easily seen that the above problem is equivalent to the continuous relaxation of Problem~\ref{prob:generalized-GM}. Clearly, this problem is a special case of the standard form~\eqref{prob:admm}. Thus, ADMM can be applied to it in a straightforward manner. The augmented Lagrangian of Problem~\ref{prob:GM-decomposed} is
\begin{multline}
L_\rho(\x_1,\x_2,\ldots,\x_D,\y) = \sum_{d=1}^{D}F^d(\x_1,\ldots,\x_d)\\  + \y^\top\left(\sum_{d=1}^D\A_d\x_d\right) + \frac{\rho}{2}\norm{\sum_{d=1}^D\A_d\x_d}^2.
\end{multline}
The $\y$ update step~\eqref{eq:y-update-general} and the computation of the residual~\eqref{eq:residual} is trivial. Let us focus on the $\x$ update step ~\eqref{eq:x-update-general}:
\begin{equation}
\x_d^{k+1} = \argmin_{\x\in \cM_d} L_\rho(\x_{[1,d-1]}^{k+1},\x,\x_{[d+1,D]}^{k},\y^k). \label{eq:x-update-v0}
\end{equation}
Denote
\begin{align}
\s_d^k &= \sum_{i=1}^{d-1} \A_i\x_i^{k+1} + \sum_{j=d+1}^{D} \A_j\x_j^{k},\\
\p_d^k &= \sum_{i=d}^{D}\cF^i\bigotimes_{j=1}^{d-1} \x_j^{k+1} \bigotimes_{l=d+1}^{i} \x_l^{k}. \quad \text{(see~\eqref{eq:bigotimes})}
\end{align}
It can be seen that (details given in the supplement)
\begin{equation}
	\sum_{i=d}^{D}F^i(\x_{[1,d-1]}^{k+1},\x,\x_{[d+1,i]}^{k}) = (\p_k)^\top\x.
\end{equation}
Thus, let $\mathrm{cst}$ be a constant independent of $\x$, we have:
\begin{multline}
L_\rho(\x_{[1,d-1]}^{k+1},\x,\x_{[d+1,D]}^{k},\y^k) = (\p_d^k)^\top\x \\ + (\y^k)^\top(\A_d\x + \s_d^k) + \frac{\rho}{2}\norm{\A_d\x + \s_d^k}^2 + \mathrm{cst},
\end{multline} 
and the subproblems~\eqref{eq:x-update-v0} are reduced to minimizing the following quadratic functions over $\cM_d$ ($d=1,2,\ldots,D$):
\begin{equation}\label{v0-qp}
\frac{1}{2}\x^\top\A_d^\top\A_d\x + \left(\A_d^\top\s_d^k + \frac{1}{\rho}(\A_d^\top\y^k + \p_d^k)\right)^\top\x.
\end{equation}

In summary, an ADGM algorithm has three main steps: 1) choose $(\A_d)_{1\le d\le D}$ and $(\cM_d)_{1\le d\le D}$ satisfying the conditions stated in Problem~\ref{prob:GM-decomposed}, 2) update $\x_d^{k+1}$ by minimizing~\eqref{v0-qp} over $\cM_d$, and 3) update $\y^{k+1}$ using~\eqref{eq:y-update-general} (and repeat 2), 3) until convergence).

\begin{comment}
\begin{table}[!t]
\centering
%\setlength\extrarowheight{10pt}
\begin{tabular}{|c|}
\hline 
General ADGM algorithm \\ \hline\hline
\begin{minipage}{0.46\textwidth}
\begin{enumerate}\addtocounter{enumi}{-1}
\item Input: $\rho, N, \epsilon, (\x_d^0)_{1\le d\le D}, \y^0$. Choose $(\A_d)_{1\le d\le D}$ and $(\cM_d)_{1\le d\le D}$ satisfying the conditions stated in Problem~\ref{prob:GM-decomposed}.
\item Repeat for $k=0,1,2,\ldots$:
\begin{enumerate}[(a)]
\item For $d=1,2,\ldots,D$, update $\x_d^{k+1}$ by minimizing~\eqref{v0-qp} over $\cM_d$.
\item Update $\y$:
\begin{equation}\label{v0-y}
\y^{k+1} =\y^{k} + \rho\left(\sum_{i=1}^D\A_i\x_i^{k+1}\right). 
\end{equation}
\item Compute the residual $r^{k+1}$ by
\begin{equation}\label{v0-r}\norm{\sum_{i=1}^{D} \A_i\x_i^{k+1}}^2 + \sum_{i=1}^{D}\norm{\A_i\x_i^{k+1} - \A_i\x_i^{k}}^2.
\end{equation}
\end{enumerate} 
Stop if $r^{k+1}\le \epsilon$ or $k \ge N$. 
\item Discretize $\x_1$ and return.
\end{enumerate} 
\end{minipage}
\\ \hline
\end{tabular}
\caption{\label{tab:algorithms}Examples of ADGM algorithms for solving the generalized graph matching problem (Problem~\ref{prob:generalized-GM}).}
\end{table}
\end{comment}

\subsection{Two simple ADGM algorithms}\label{sec:two-simple-adgm}
Let us follow the above three steps with two examples.

\noindent\textbf{Step 1: Choose $(\A_d)_{1\le d\le D}$ and $(\cM_d)_{1\le d\le D}$.} First, $(\cM_d)_{1\le d\le D}$ take values in one of the following two sets:
\begin{align}
\Mr &= \small\set{\x: \mbox{ sum of each row of } \mat(\x) \mbox{ is } \le 1},\label{eq:mr}\\
\Mc &= \small\set{\x: \mbox{ sum of each column of } \mat(\x) \mbox{ is } \le 1},\label{eq:mc}
\end{align}
such that \emph{both $\Mr$ and $\Mc$ are taken at least once}. If no occlusion is allowed in $\cG_1$ (respectively $\cG_2$), then the term ``$\le 1$'' is replaced by ``$=1$'' for $\cM_\mathrm{r}$ (respectively $\cM_\mathrm{c}$). If many-to-many matching is allowed, then these inequality constraints are removed. In either case, $\Mr$ and $\Mc$ are closed and convex. Clearly, since $\Mr\cap\Mc = \cM$, condition~\eqref{eq:v0-set-condition} is satisfied.\\
Second, to impose $\x_1=\x_2=\cdots = \x_D$, we can for example choose $(\A_d)_{1\le d\le D}$ such that
\begin{equation}\label{equa1}
\x_1=\x_2,\quad \x_1=\x_3,\ldots,\quad\x_1=\x_D
\end{equation}
or alternatively
\begin{equation}\label{equa2}
\x_1=\x_2,\quad \x_2=\x_3,\ldots,\quad\x_{D-1}=\x_D.
\end{equation}
It is easily seen that the above two sets of constraints can be both expressed under the general form~\eqref{eq:v0-equal}. Each choice leads to a different algorithm. Let ADGM1 denote the one obtained from~\eqref{equa1} and ADGM2 obtained from~\eqref{equa2}.

\noindent\textbf{Step 2: Update $\x_d^{k+1}$.} Plugging~\eqref{equa1} and~\eqref{equa2} into~\eqref{eq:v0-equal}, the subproblems~\eqref{v0-qp} are reduced to (details given in the supplement) 
\begin{equation}\label{eq:v0-x-update}
\x_d^{k+1} = \argmin_{\x\in \cM_d} \set{\frac{1}{2}\norm{\x}^2 - \c_d^\top \x},
\end{equation}
where $(\c_d)_{1\le d\le D}$ are defined as follows, for ADGM1:
\begin{align}
\c_1&= \frac{1}{D-1}\left(\sum_{d=2}^{D}\x_d^k - \frac{1}{\rho}\sum_{d=2}^{D} \y_d^k - \frac{1}{\rho} \sum_{d=1}^{D} \cF^d\bigotimes_{i=2}^d\x_i^k\right), \label{eq:v1-c1}\\
\c_d&= \x_1^{k+1} +\frac{1}{\rho}\y_d^k - \frac{1}{\rho}\left(\sum_{i=d}^{D} \cF^i\bigotimes_{i=1}^{d-1} \x_i^{k+1} \bigotimes_{j=d+1}^i\x_{j}^k\right) \label{eq:v1-cd}
\end{align}
for $2\le d\le D$, and for ADGM2:
\begin{align}
\c_1 &= \x_2^k - \frac{1}{\rho}\y_2^k -\frac{1}{\rho}\sum_{d=1}^{D} \cF^d\bigotimes_{i=2}^d \x_i^k, \label{eq:v2-c1}\\
\c_D &= \x_{D-1}^{k+1} + \frac{1}{\rho}\y_D^k - \frac{1}{\rho}\cF^D\bigotimes_{i=1}^{D-1} \x_i^{k+1}, \label{eq:v2-cD} \\
\c_d &= \frac{1}{2}(\x_{d-1}^{k+1}+\x_{d+1}^k) + \frac{1}{2\rho}(\y_d^k - \y_{d+1}^k) \notag\\ &- \frac{1}{2\rho}\sum_{i=d}^{D} \cF^i\bigotimes_{i=1}^{d-1} \x_i^{k+1} \bigotimes_{j=d+1}^i\x_{j}^k\label{eq:v2-cd}
\end{align}
for $2\le d\le D-1$.

\noindent\textbf{Step 3: Update $\y^{k+1}$.} From~\eqref{equa1} and~\eqref{equa2}, it is seen that this step is reduced to $\y_d^{k+1} = \y_d^k + \rho(\x_{1}^{k+1} - \x_d^{k+1})$ for ADGM1 and $\y_d^{k+1} = \y_d^k + \rho(\x_{d-1}^{k+1} - \x_d^{k+1})$ for ADGM2. 

\noindent\textit{Remark.} When $D=2$ the two algorithms are identical. 

Note that~\eqref{eq:v0-x-update} means $\x_d^{k+1}$ is the projection of $\c_d$ onto $\cM_d$. Since $(\cM_d)_{1\le d\le D}$ obey only row-wise or column-wise constraints, the projection becomes row-wise or column-wise and can be solved based on the projection onto a simplex~\cite{condat2014fast}. We show how to do that and give sketches of the above algorithms in the supplement.

\subsection{Convergent ADGM}\label{sec:convergent-adgm}
Note that the objective function in Problem~\ref{prob:GM-decomposed} is neither separable nor convex in general. Convergence of ADMM for this type of functions is unknown. Indeed, our ADGM algorithms do not always converge, especially for small values of the penalty parameter $\rho$. When $\rho$ is large, they are likely to converge. However, we also notice that small $\rho$ often (but not always) achieves better objective values.  Motivated by these observations, we propose the following adaptive scheme that we find to work very well in practice: starting from a small initial value $\rho_0$, the algorithm runs for $T_1$ iterations to stabilize, after that, if no improvement of the residual $r^k$ is made every $T_2$ iterations, then we increase $\rho$ by a factor $\beta$ and continue. The intuition behind this scheme is simple: we hope to reach a good objective value with a small $\rho$, but if this leads to slow (or no) convergence, then we increase $\rho$ to put more penalty on the consensus of the variables and that would result in faster convergence. Using this scheme, we observe that our algorithms always converge in practice. In the experiments, we set $T_1 = 300, T_2 = 50, \beta = 2$ and $\rho_0 = \frac{n}{1000}$.

\begin{comment}
\begin{figure}
\centering
\includegraphics[width=0.8\linewidth]{residuals}
\caption{\label{fig:residuals}The residual $r^k$ per iteration of ADGM. Adaptive parameters: $T_1 = 20, T_2 = 5, \beta = \rho_0 = 2.0$ (\cf Section~\ref{sec:convergent-adgm}). Run on a third-order Motorbike matching with 15 outliers (\cf Section~\ref{sec:cars-motorbikes} for data and model description).}
\end{figure}
\end{comment}

\section{Experiments}\label{sec:experiments}
We adopt the adaptive scheme in Section~\ref{sec:convergent-adgm} to two ADGM algorithms presented in Section~\ref{sec:two-simple-adgm}, and denote them respectively ADGM1 and ADGM2. In pairwise settings, however, since these two algorithms are identical, we denote them simply ADGM. We compare ADGM and ADGM1/ADGM2 to the following state of the art methods:

\noindent \textbf{Pairwise:} Spectral Matching with Affine Constraint (SMAC)~\cite{cour2007balanced}, Integer Projected Fixed Point (IPFP)~\cite{leordeanu2009integer}, Reweighted Random Walk Matching (RRWM)~\cite{cho2010reweighted}, Dual Decomposition with Branch and Bound (DD)~\cite{torresani2013dual} and Max-Pooling Matching (MPM)~\cite{cho2014finding}. We should note that DD is only used in the experiments using the same energy models presented in~\cite{torresani2013dual}. For the other experiments, DD is excluded due to the prohibitive execution time. Also, as suggested in~\cite{leordeanu2009integer}, we use the solution returned by Spectral Matching (SM)~\cite{leordeanu2005spectral} as initialization for IPFP.

\noindent \textbf{Higher-order:} Probabilistic Graph Matching (PGM)~\cite{zass2008probabilistic}, Tensor Matching (TM)~\cite{duchenne2011tensor},  Reweighted Random Walk Hypergraph Matching (RRWHM)~\cite{lee2011hyper} and Block Coordinate Ascent Graph Matching (BCAGM)~\cite{nguyen2015flexible}. For BCAGM, we use MPM~\cite{cho2014finding} as subroutine because it was shown in~\cite{nguyen2015flexible} (and again by our experiments) that this variant of BCAGM (denoted by ``BCAGM+MP'' in~\cite{nguyen2015flexible}) outperforms the other variants thanks to the effectiveness of MPM. Since there is no ambiguity, in the sequel we denote this variant ``BCAGM'' for short.

\begin{table*}[!htb]
\begin{minipage}[b]{0.36\linewidth}\small
\centering
	\begin{tabular}{|c|l|ccc|}\hline
	 &  Methods & Error 	& Global   & Time \\
		&		    &	(\%) 	& opt. (\%)			& (s) \\	\hline\hline
	\multirow{6}{*}{\begin{turn}{90}House\end{turn}}  
	      	&MPM	& 42.32			& 0					& 0.02 \\
			 &RRWM	& 90.51			& 0					& 0.01 \\
	      	&IPFP	& 87.30			& 0					& 0.02 \\
	      	&SMAC	& 81.11			& 0					& 0.18 \\
			 &DD	& \textbf{0}	& \textbf{100}		& 14.20 \\
	      	&ADGM	& \textbf{0}	&\textbf{100}		& 0.03 \\\hline\hline
	\multirow{6}{*}{\begin{turn}{90}Hotel\end{turn}}    
	      	&MPM	& 21.49			& 44.80				& 0.02 \\
			 &RRWM	& 85.05			& 0					& 0.01 \\
	      	&IPFP	& 85.37			& 0					& 0.02 \\
	      	&SMAC	& 71.33			& 0					& 0.18 \\
			 &DD	& \textbf{0.19}	& \textbf{100}		& 13.57 \\
	      	&ADGM	& \textbf{0.19}	& \textbf{100}		& 0.02 \\\hline
	  \end{tabular}  
	    \caption{Results on House and Hotel sequences using the \textbf{pairwise model A}, described in Section~\ref{sec:house-hotel} and previously proposed in~\cite{torresani2013dual}.}
	    \label{tab:result-torresani}
\end{minipage}\hfill
\begin{minipage}[b]{0.62\linewidth}
    \begin{subfigure}[t]{0.47\linewidth}
        \centering
        \includegraphics[width=\linewidth]{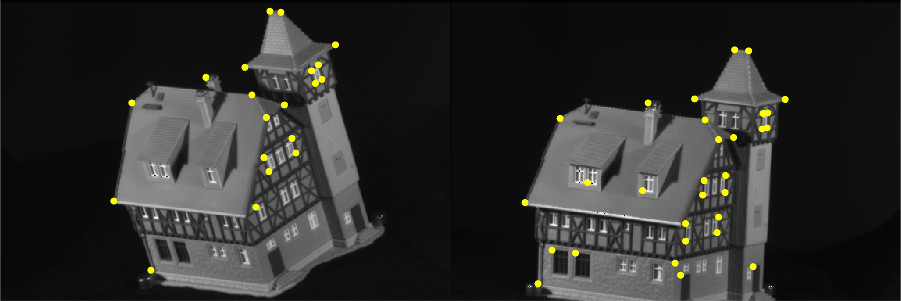}\vspace{-5pt}
        \caption{\small 20 pts vs 30 pts (10 outliers)}
    \end{subfigure}%
    ~
    \begin{subfigure}[t]{0.47\linewidth}
        \centering
        \includegraphics[width=\linewidth]{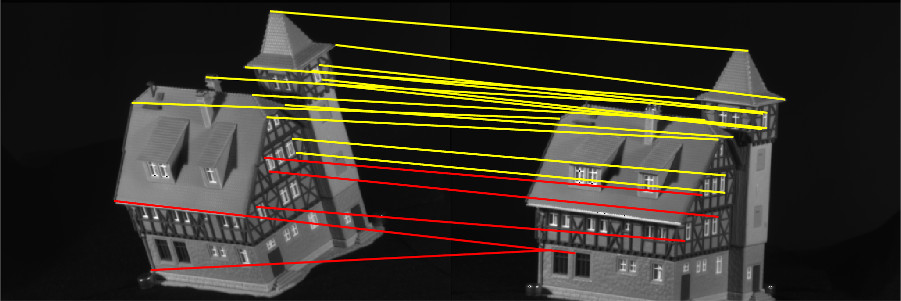}\vspace{-5pt}
        \caption{\small MPM 15/20 (352.4927)}
    \end{subfigure}\\%
    \begin{subfigure}[t]{0.47\linewidth}
        \centering
        \includegraphics[width=\linewidth]{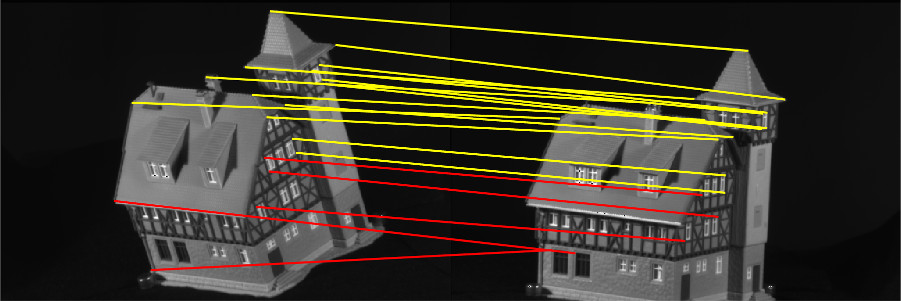}\vspace{-5pt}
        \caption{\small RRWM 15/20 (352.4927)}
    \end{subfigure}%
    ~
    \begin{subfigure}[t]{0.47\linewidth}
        \centering
        \includegraphics[width=\linewidth]{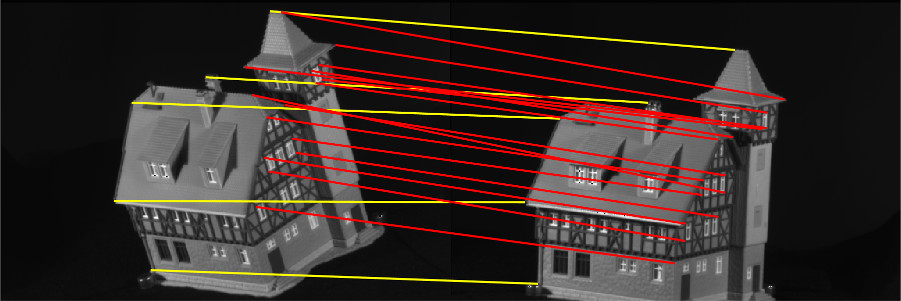}\vspace{-5pt}
        \caption{\small IPFP 5/20 (316.9299)}
    \end{subfigure}\\%
    \begin{subfigure}[t]{0.47\linewidth}
        \centering
        \includegraphics[width=\linewidth]{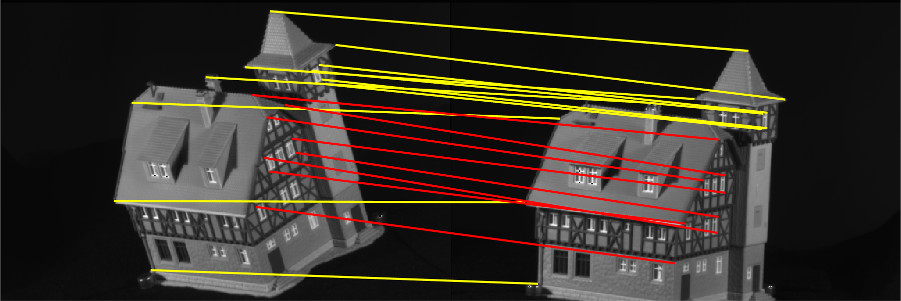}\vspace{-5pt}
        \caption{\small SMAC 12/20 (315.0426)}
    \end{subfigure}%
    ~
    \begin{subfigure}[t]{0.47\linewidth}
        \centering
        \includegraphics[width=\linewidth]{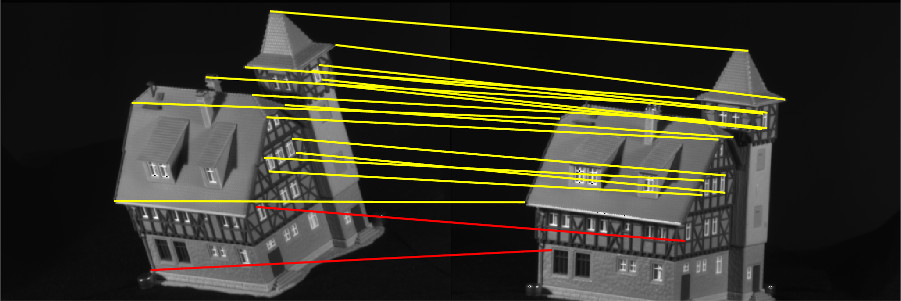}\vspace{-5pt}
        \caption{\small ADGM \textbf{18/20} (\textbf{353.3569})}
    \end{subfigure}%
    \captionof{figure}{\label{fig:2nd-demo-Houses}House matching using the \textbf{pairwise model B} described in Section~\ref{sec:house-hotel}. Ground-truth value is $343.1515$. (Best viewed in color.)}
\end{minipage}
\end{table*}

We should note that, while we formulated the graph matching as a \emph{minimization} problem, most of the above listed methods are \emph{maximization} solvers and many models/objective functions in previous work were designed to be maximized. For ease of comparison, ADGM is also converted to a maximization solver (by letting it minimize the additive inverse of the objective function), and the results reported in this section are for the maximization settings (\ie higher objective values are better). In the experiments, we also use some pairwise minimization models (such as the one from~\cite{torresani2013dual}), which we convert to maximization problems as follows: after building the affinity matrix $\M$ from the (minimization) potentials, the new (maximization) affinity matrix is computed by $\max(\M) - \M$ where $\max(\M)$ denotes the greatest element of $\M$. Note that one cannot simply take $-\M$ because some of the methods only work for non-negative potentials.

And last, due to space constraints, we leave the reported running time for each algorithm in the supplement (except for the very first experiment where this can be presented compactly). In short, ADGM is faster than SMAC~\cite{cour2007balanced} (in pairwise settings) and ADGM1/ADGM2 are faster than TM~\cite{duchenne2011tensor} (in higher-order settings) while being slower than the other methods.

\subsection{House and Hotel}\label{sec:house-hotel}
The CMU House and Hotel sequences\footnote{\scriptsize\url{http://vasc.ri.cmu.edu/idb/html/motion/index.html}} have been widely used in previous work to evaluate graph matching algorithms. It consists of 111 frames of a synthetic house and 101 frames of a synthetic hotel. Each frame in these sequences is manually labeled with 30 feature points.

\noindent\textbf{Pairwise model A.} In this experiment we match all possible pairs of images in each sequence, with all $30$ points (\ie no outlier). A Delaunay triangulation is performed for these $30$ points to obtain the graph edges. The unary terms are the distances between the Shape Context descriptors~\cite{belongie2002shape}. The pairwise terms when matching $(i_1,j_1)$ to $(i_2,j_2)$ are
\begin{equation}\label{eq:pairwise-tkr}
\cF_{ij}^2 = \eta \exp\left(\delta^2/\sigma_l^2\right) + (1- \eta) \exp\left(\alpha^2/\sigma_a^2\right) - 1
\end{equation} 
where $\eta,\sigma_l,\sigma_a$ are some weight and scaling constants and $\delta,\alpha$ are computed from $d_1 = \|\overrightarrow{i_1j_1}\|$ and $d_2 = \|\overrightarrow{i_2j_2}\|$ as
\begin{equation}
\delta = \frac{\abs{d_1 - d_2}}{
d_1+d_2},\quad
\alpha = \arccos\left(\frac{\overrightarrow{i_1j_1}}{d_1} \cdot \frac{\overrightarrow{i_2j_2}}{d_2}\right).
\end{equation}
This experiment is reproduced from~\cite{torresani2013dual} using their energy model files\footnote{\scriptsize\url{http://www.cs.dartmouth.edu/~lorenzo/Papers/tkr\_pami13\_data.zip}.}. It should be noted that in~\cite{torresani2013dual}, the above unary potentials are subtracted by a large number to prevent occlusion. We refer the reader to~\cite{torresani2013dual} for further details.  For ease of comparison with the results reported in~\cite{torresani2013dual}, here we also report the performance of each algorithm in terms of overall percentage of mismatches and frequency of reaching the global optimum. Results are given in Table~\ref{tab:result-torresani}. One can observe that DD and ADGM always reached the global optima, but ADGM did it hundreds times faster. Even the recent methods RRWM and MPM performed poorly on this model (only MPM produced acceptable results). Also, we notice a dramatic decrease in performance of SMAC and IPFP compared to the results reported in~\cite{torresani2013dual}. We should note that the above potentials, containing both positive and negative values, are defined for a \emph{minimization} problem. It was unclear how those \emph{maximization} solvers were used in~\cite{torresani2013dual}. For the reader to be able to reproduce the results, we make our software publicly available.

\begin{figure*}[!htb]
\centering
    \begin{subfigure}[t]{0.24\textwidth}
        \centering
        \includegraphics[width=\linewidth]{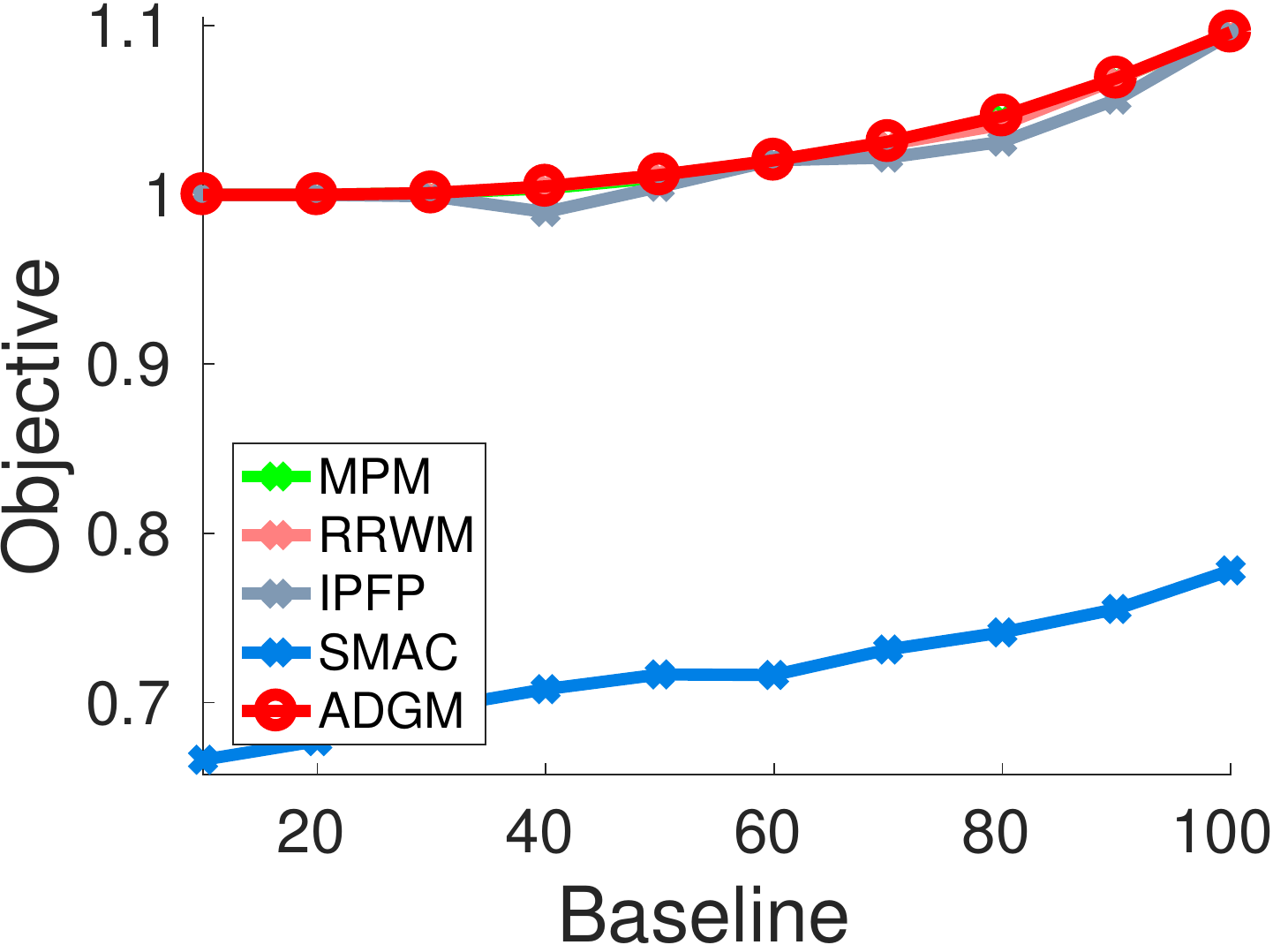}\\
        \includegraphics[width=\linewidth]{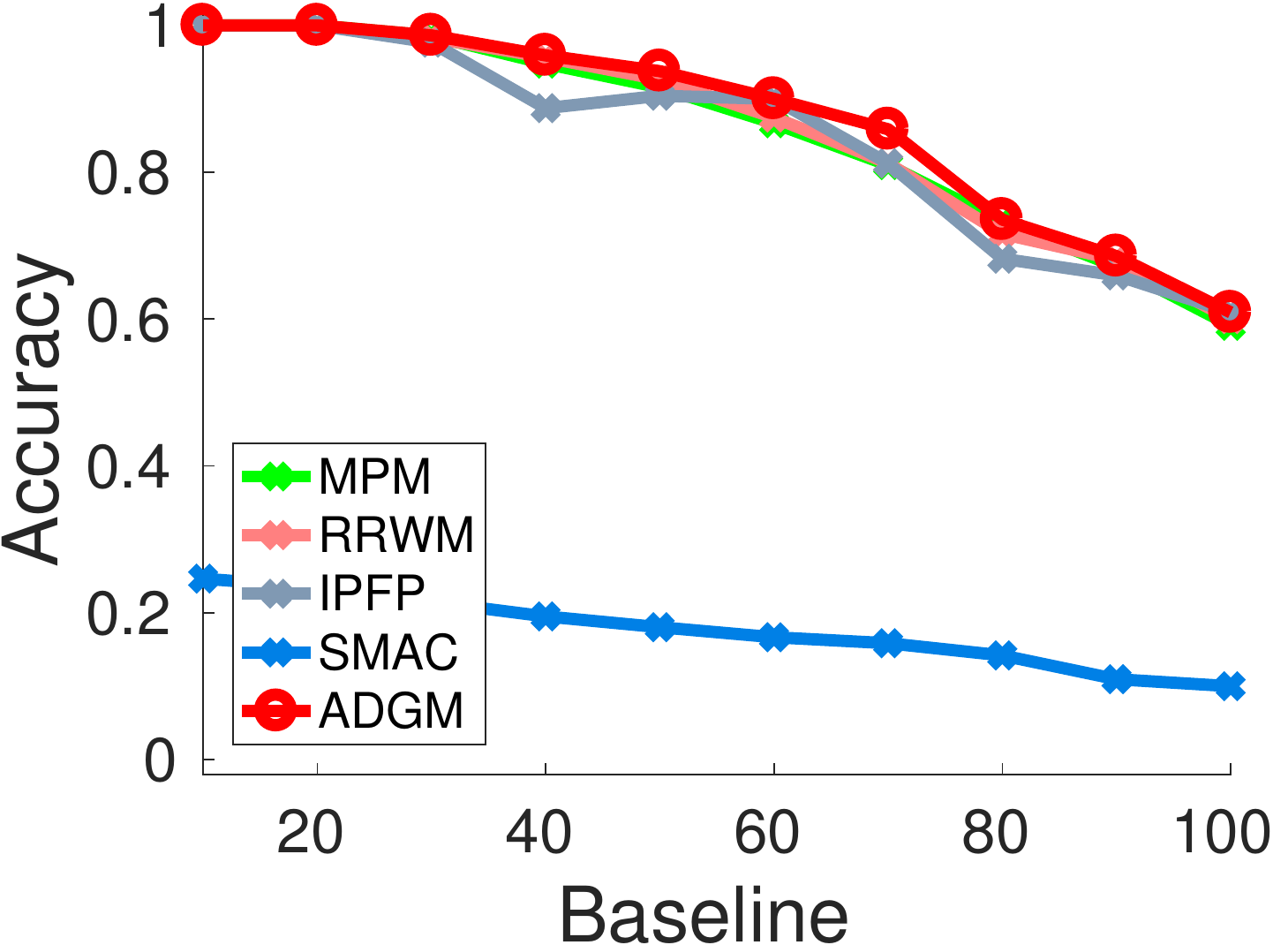}
        \caption{House: 20 pts vs 30 pts}
    \end{subfigure}%
    ~ 
    \begin{subfigure}[t]{0.24\textwidth}
            \centering
            \includegraphics[width=\linewidth]{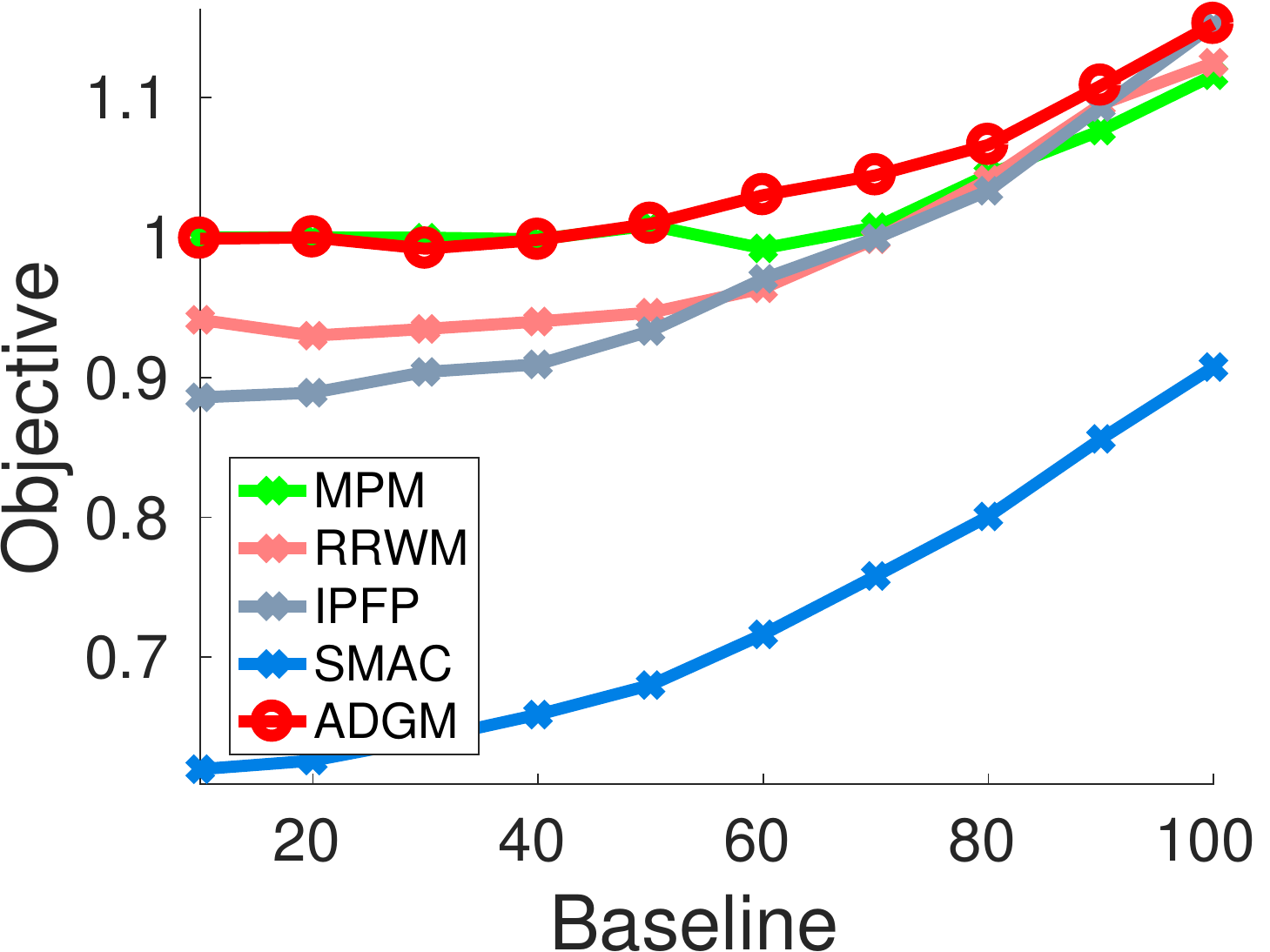}\\
            \includegraphics[width=\linewidth]{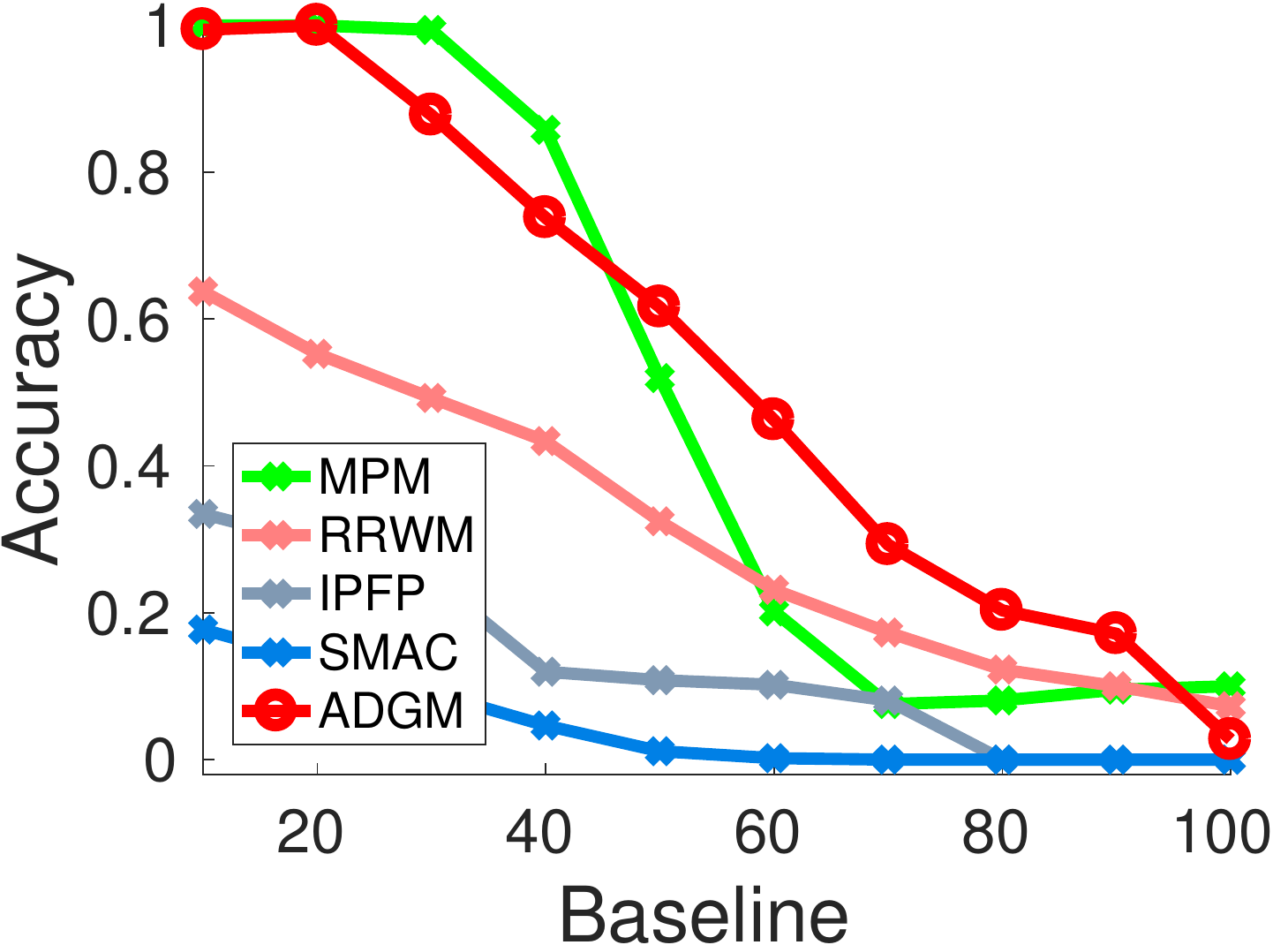}
            \caption{House: 10 pts vs 30 pts}
    \end{subfigure}%
    ~
    \begin{subfigure}[t]{0.24\textwidth}
        \centering
        \includegraphics[width=\linewidth]{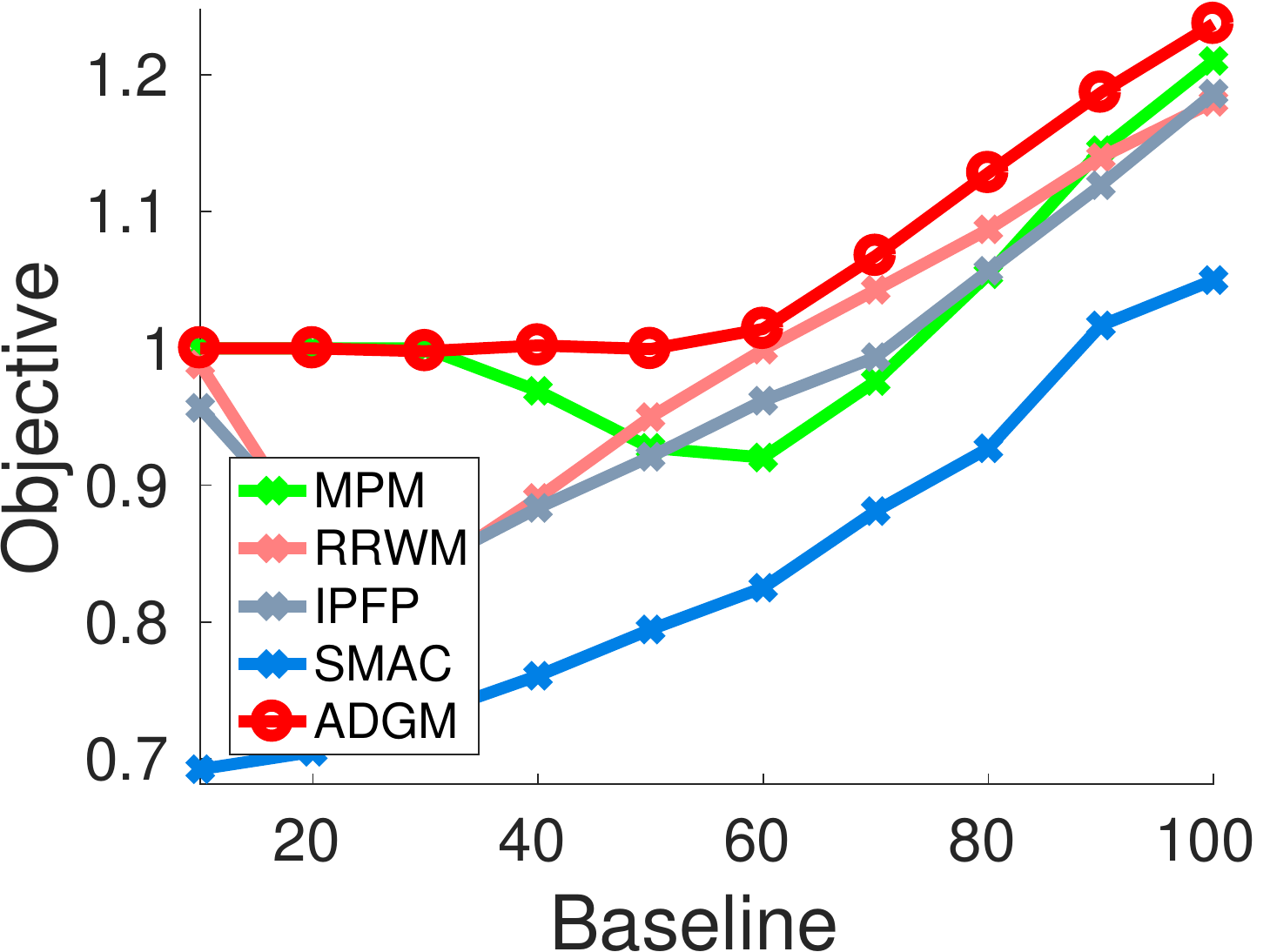}\\
        \includegraphics[width=\linewidth]{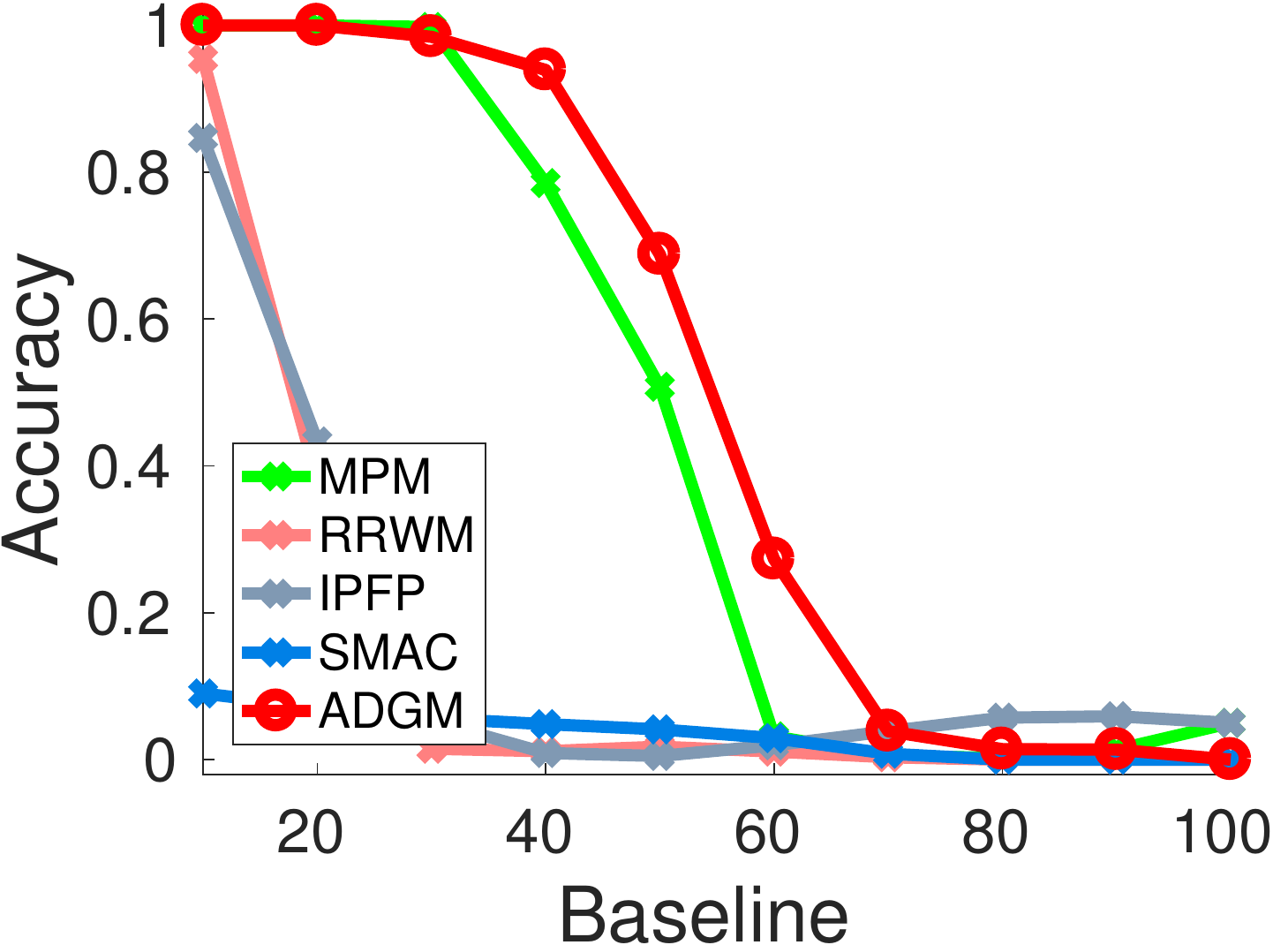}
        \caption{Hotel: 20 pts vs 30 pts}
    \end{subfigure}%
    ~ 
    \begin{subfigure}[t]{0.24\textwidth}
        \centering
        \includegraphics[width=\linewidth]{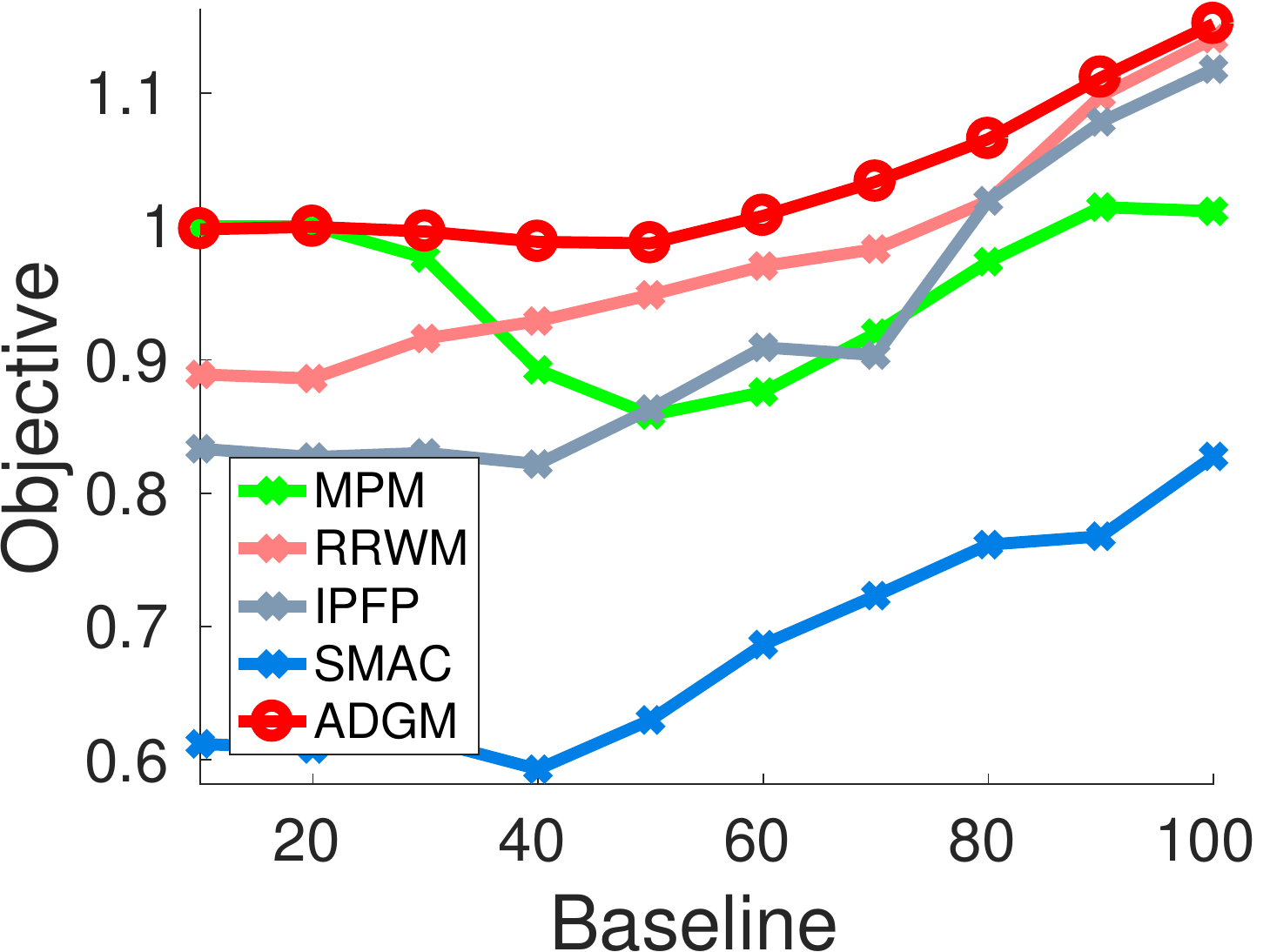}\\
        \includegraphics[width=\linewidth]{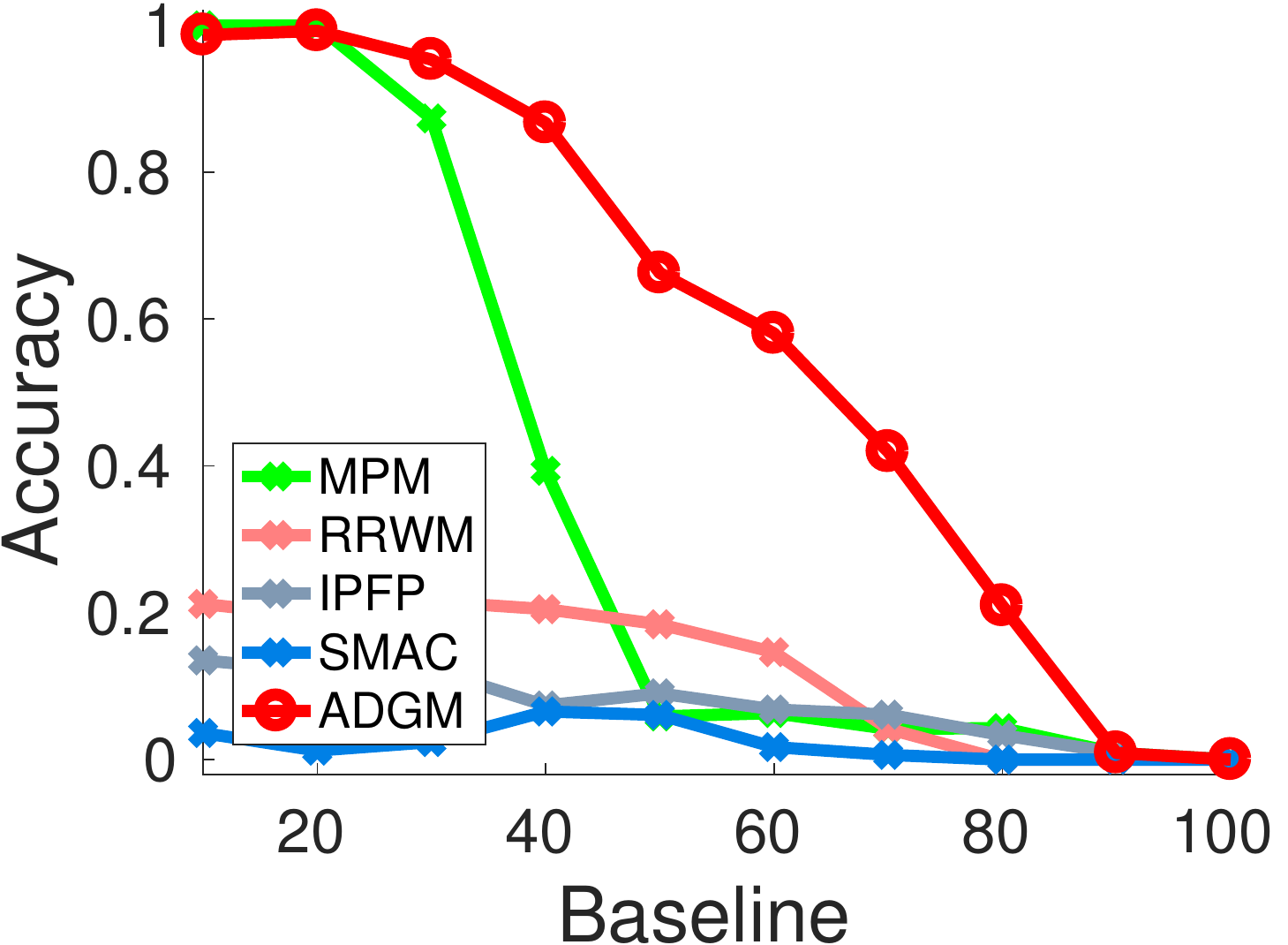}
        \caption{Hotel: 10 pts vs 30 pts}
    \end{subfigure}%
    \caption{\label{fig:results-2nd-house}Results on House and Hotel sequences using the \textbf{pairwise model B} described in Section~\ref{sec:house-hotel}.}
\end{figure*}

\noindent\textbf{Pairwise model B.} In this experiment, we match all possible pairs of the sequence with the baseline (\ie the separation between frames, \eg the baseline between frame $5$ and frame $105$ is $100$) varying from $10$ to $100$ by intervals of $10$. For each pair, we match $10,20$ and $30$ points in the first image to $30$ points in the second image. We set the unary terms to $0$ and compute the pairwise terms as
\begin{equation}
\cF_{ij}^2 = \exp\left(- \abs{\|\overrightarrow{i_1j_1}\| - \|\overrightarrow{i_2j_2}\|} / \sigma^2\right),
\end{equation} where $\sigma^2 = 2500$. It should be noted that the above pairwise terms are computed for every pair $(i_1,j_1)$ and $(i_2,j_2)$, \ie the graphs are fully connected. This experiment has been performed on the House sequence in previous work, including~\cite{cho2010reweighted} and~\cite{nguyen2015flexible}. Here we consider the Hotel sequence as well.
We report the average objective ratio (which is the ratio of the obtained objective value over the ground-truth value) and the average accuracy for each algorithm in Figure~\ref{fig:results-2nd-house}.  Due to space constraints, we only show the results for the harder cases where occlusion is allowed, and leave the other results in the supplement. As one can observe, ADGM produced the highest objective values in almost all the tests.

\noindent\textbf{Third-order model.} This experiment has the same settings as the previous one, but uses a third-order model proposed in~\cite{duchenne2011tensor}. We set the unary and pairwise terms to $0$ and compute the potentials when matching two triples of points $(i_1,j_1,k_1)$ and $(i_2,j_2,k_2)$ as
\begin{equation}
\cF^3_{ijk} = \exp\left(-\norm{f_{i_1j_1k_1} - f_{i_2j_2k_2}}^2/\gamma\right),
\end{equation}
where $f_{ijk}$ is a feature vector composed of the angles of the triangle $(i,j,k)$, and $\gamma$ is the mean of all squared distances. We report the results for House sequence in Figure~\ref{fig:results-3rd-house} and provide the other results in the supplement. One can observe that ADGM1 and ADGM2 achieved quite similar performance, both were competitive with BCAGM~\cite{nguyen2015flexible} while outperformed all the other methods.

\begin{figure}[!htb]
\centering
    \begin{subfigure}[t]{0.48\linewidth}
        \centering
        \includegraphics[width=\linewidth]{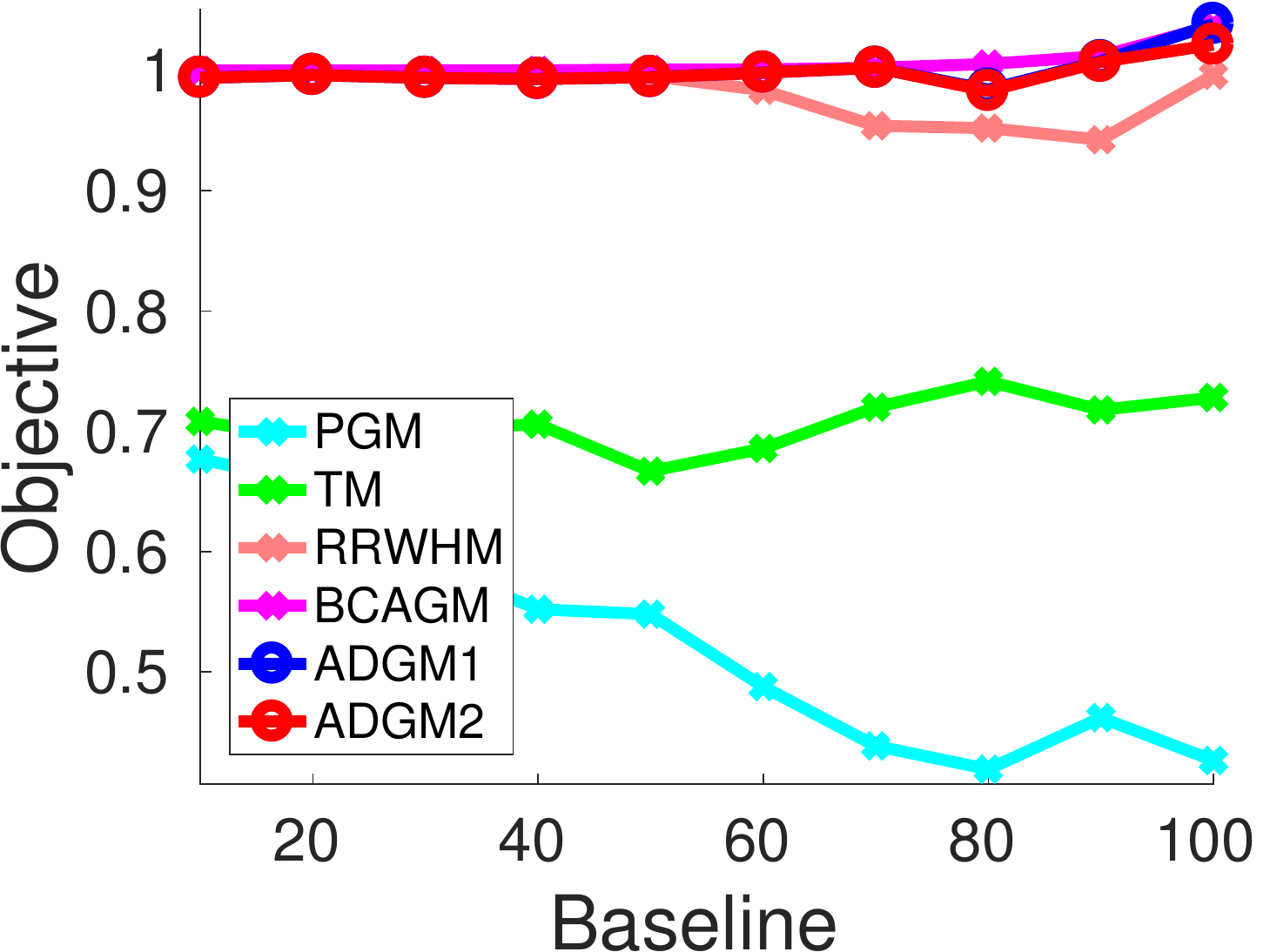}\\
        \includegraphics[width=\linewidth]{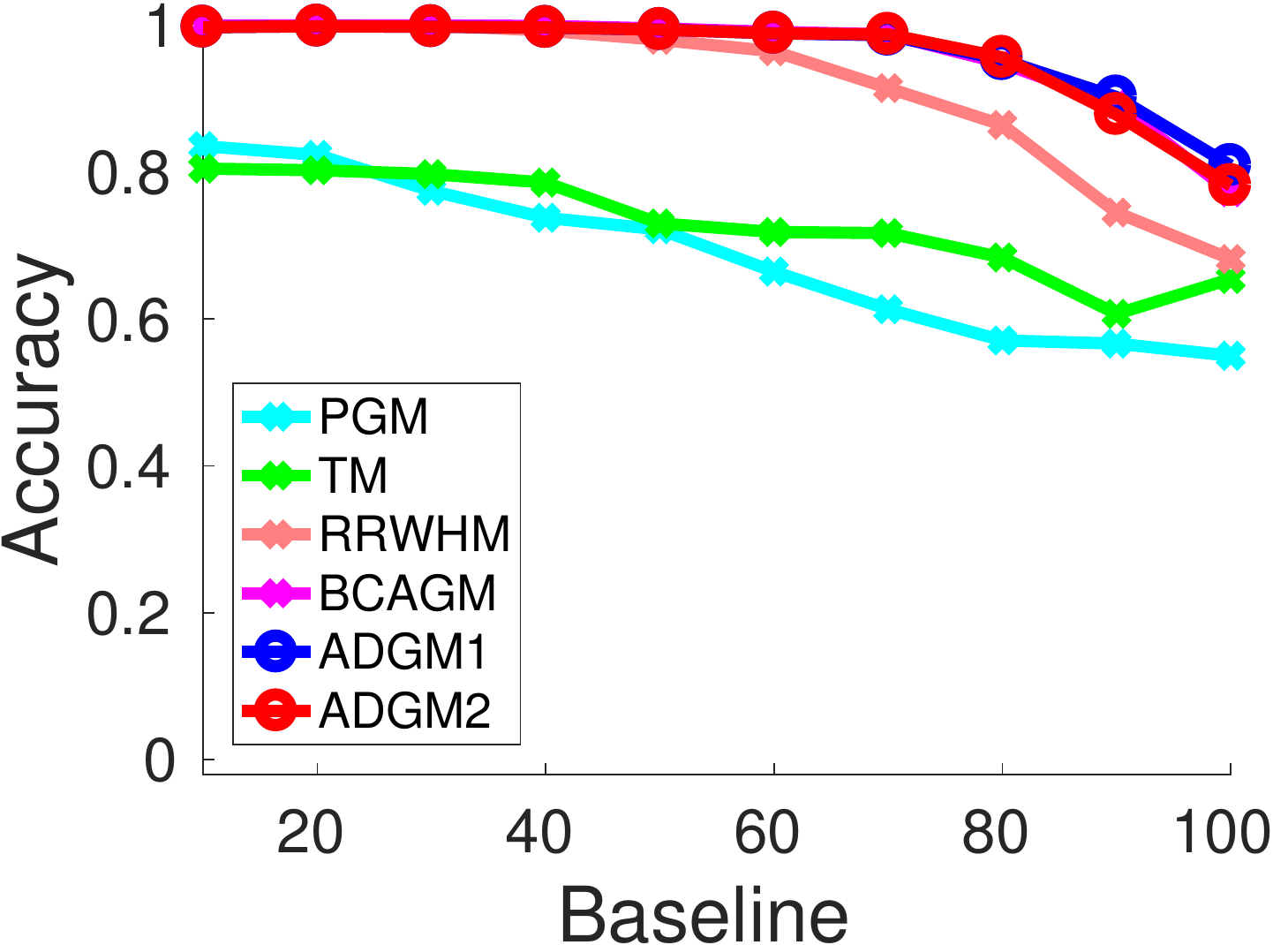}
        \caption{20 pts vs 30 pts}
    \end{subfigure}%
    ~ 
    \begin{subfigure}[t]{0.48\linewidth}
            \centering
            \includegraphics[width=\linewidth]{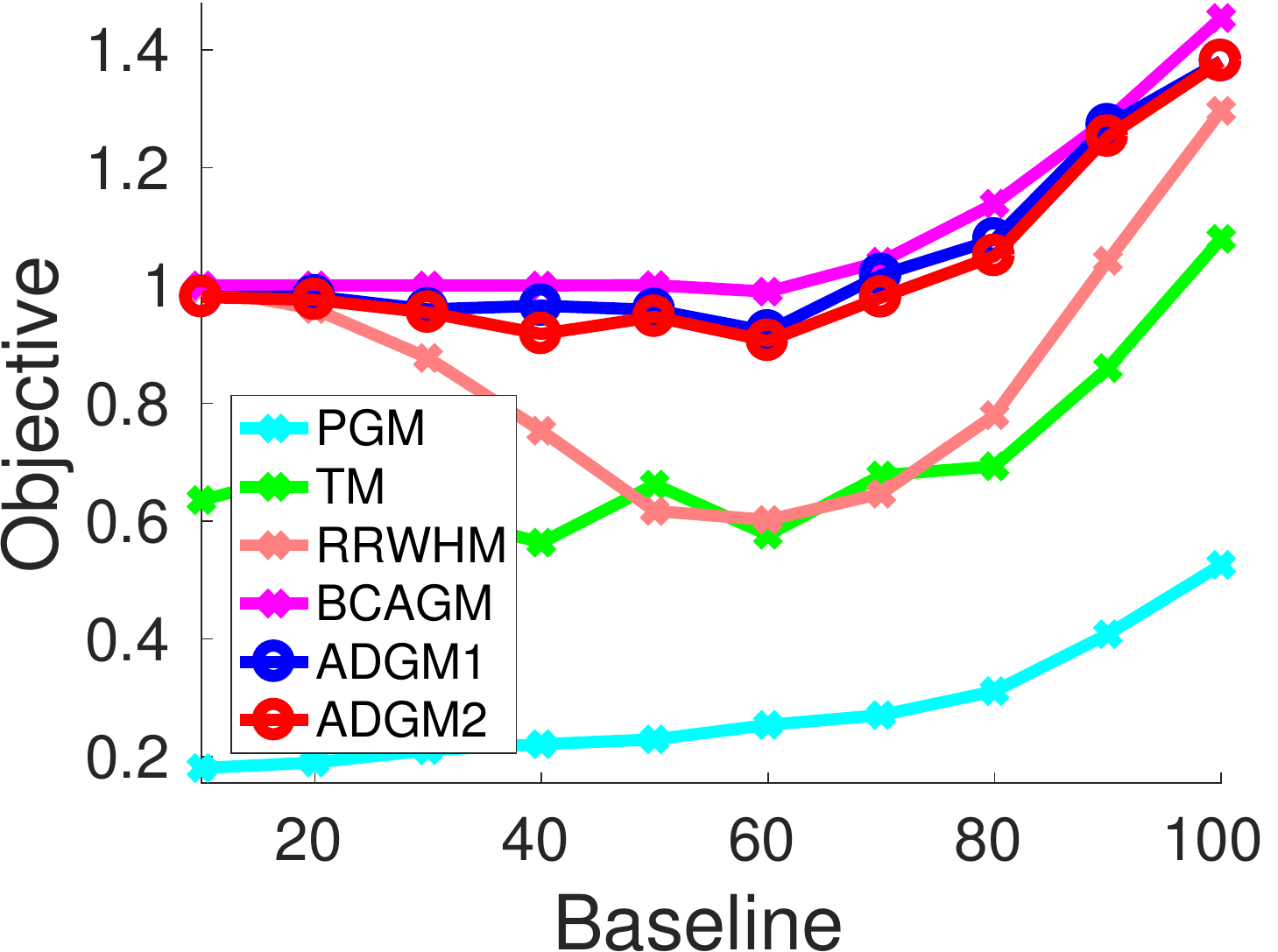}\\
            \includegraphics[width=\linewidth]{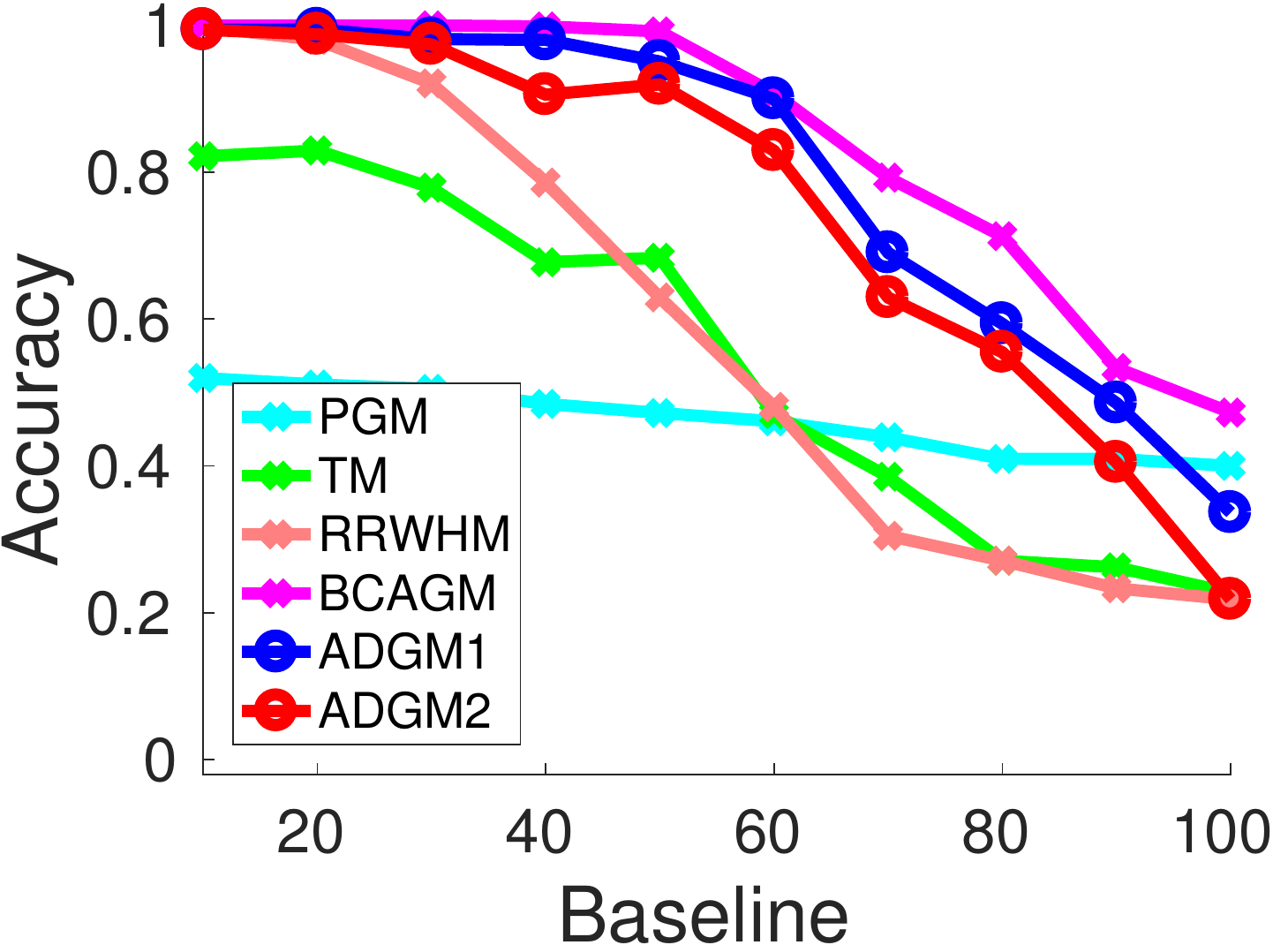}
            \caption{10 pts vs 30 pts}
    \end{subfigure}%
    \caption{\label{fig:results-3rd-house}Results on House sequence using the \textbf{third-order model} described in Section~\ref{sec:house-hotel}.}
\end{figure}

\subsection{Cars and Motorbikes}\label{sec:cars-motorbikes}
The Cars and Motorbikes dataset was introduced in~\cite{leordeanu2012unsupervised} and has been used in previous work for evaluating graph matching algorithms. It consists of $30$ pairs of car images and $20$ pairs of motorbike images with different shapes, view-points and scales. Each pair contains both inliers (chosen manually) and outliers (chosen randomly).

\begin{figure}[!tb]
\centering
    \begin{subfigure}[t]{0.48\linewidth}
        \centering
        \includegraphics[width=\linewidth]{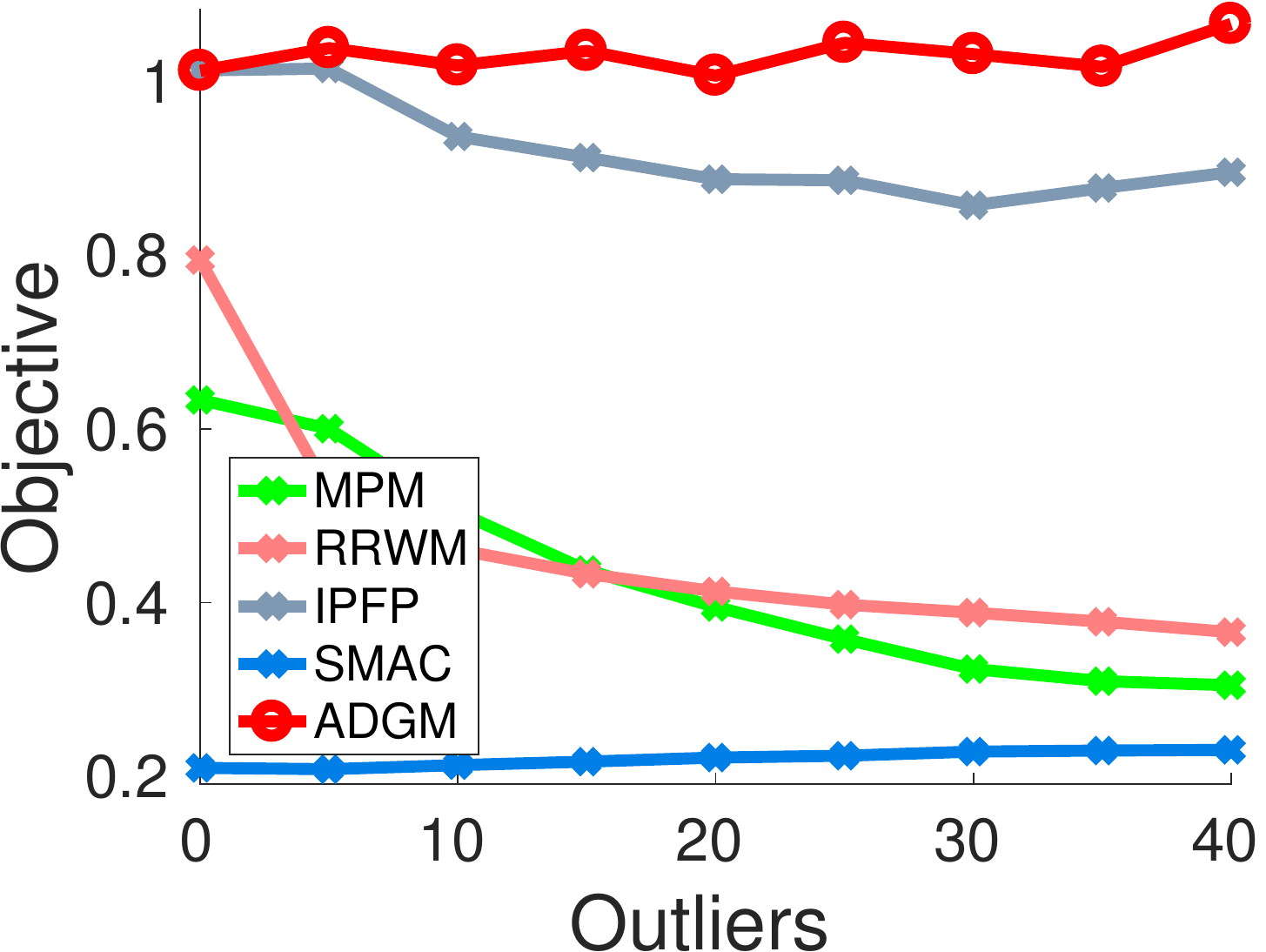}\\
        \includegraphics[width=\linewidth]{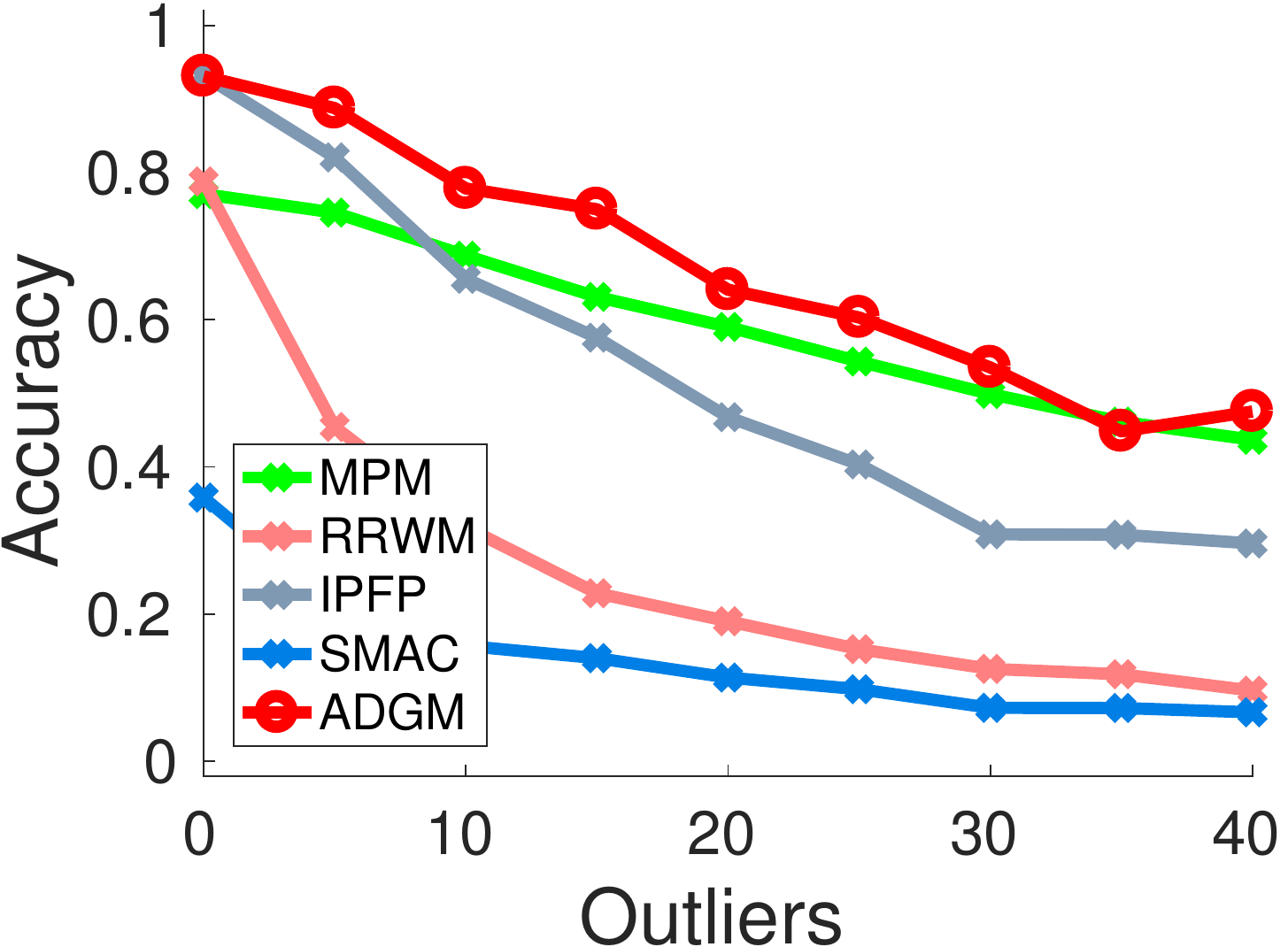}
        \caption{Cars}
    \end{subfigure}%
    ~
    \begin{subfigure}[t]{0.48\linewidth}
        \centering
        \includegraphics[width=\linewidth]{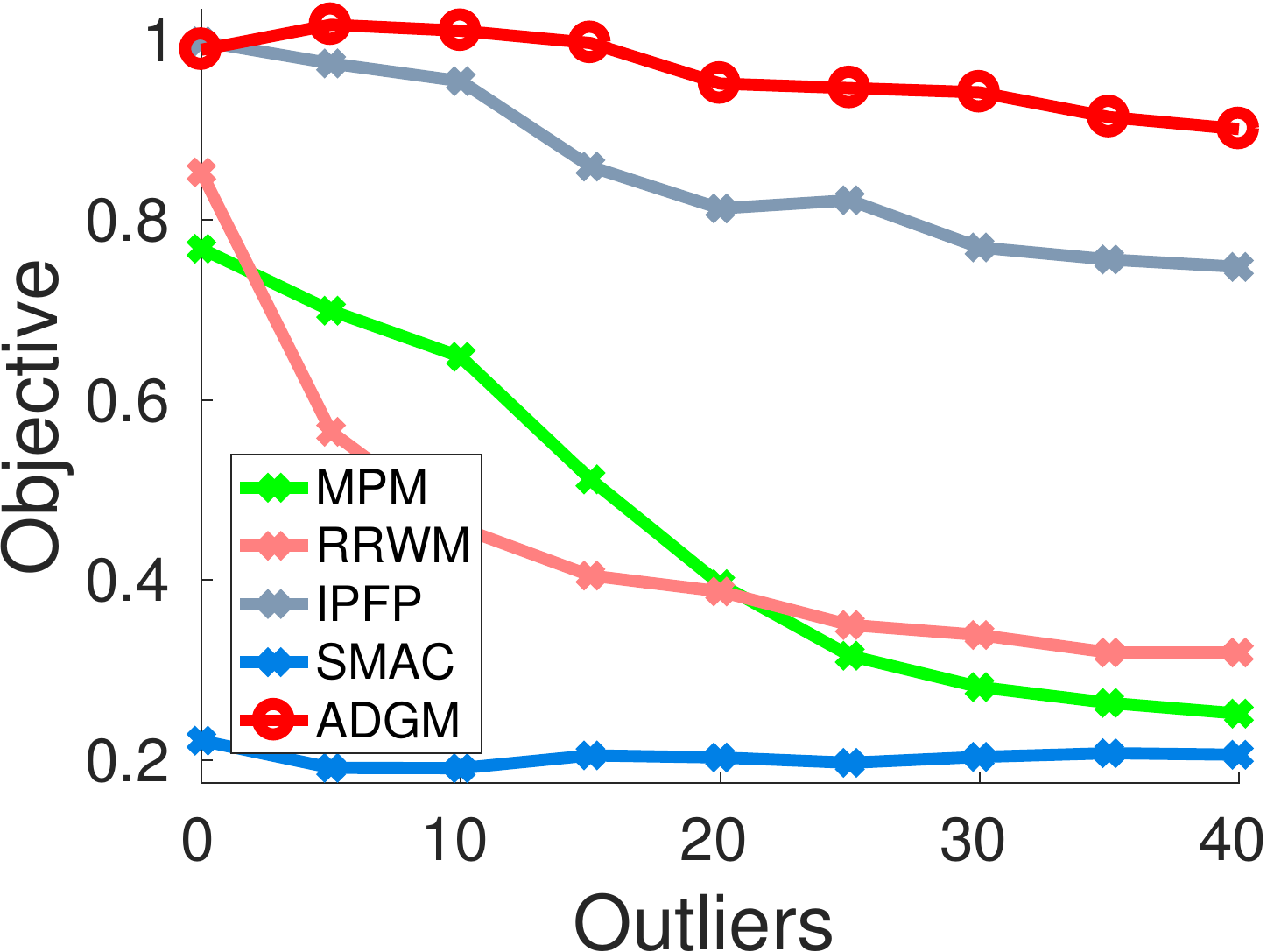}
        \includegraphics[width=\linewidth]{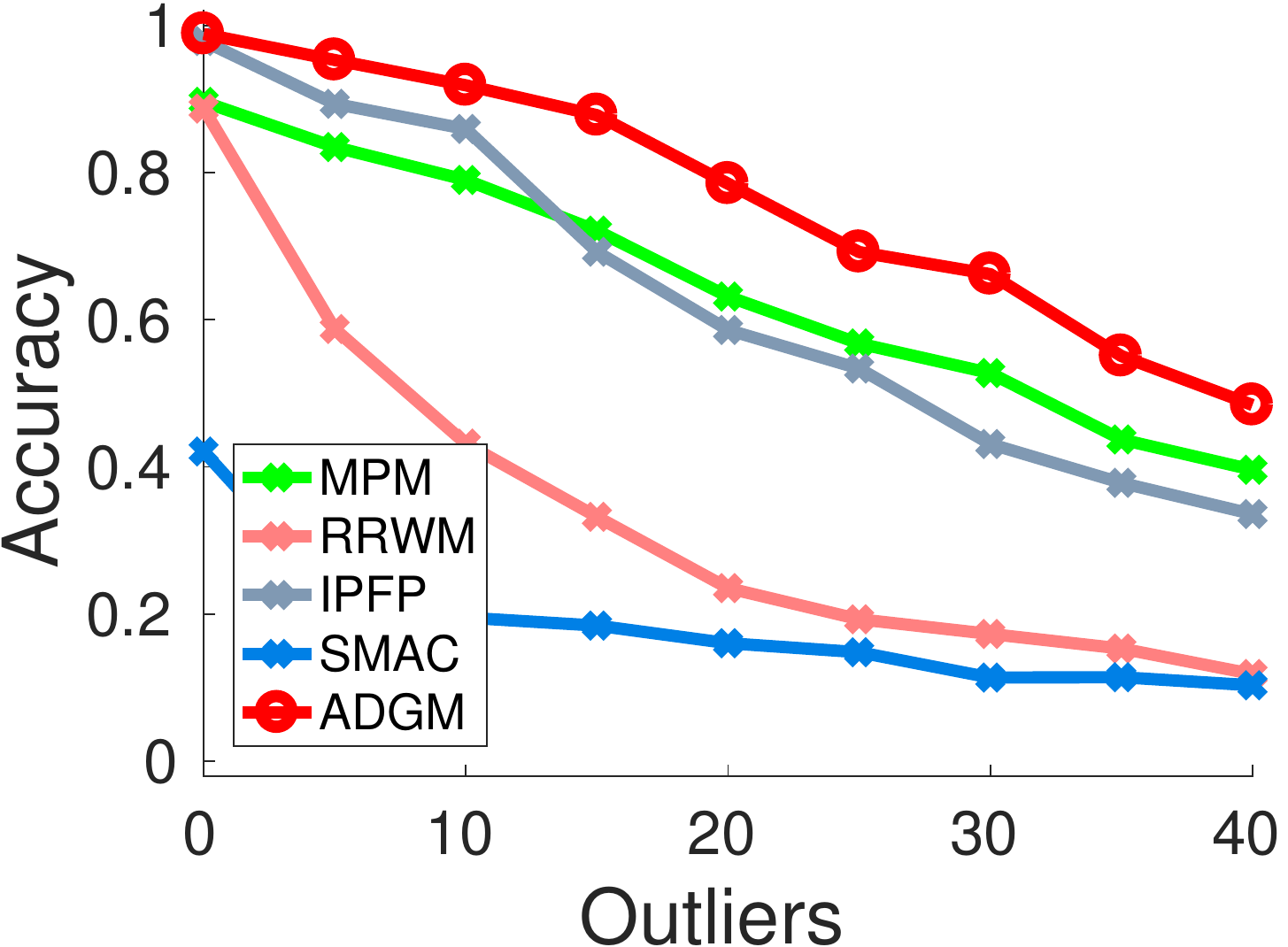}
        \caption{Motorbikes}
    \end{subfigure}%
    \caption{\label{fig:results-2nd-CarsBikes}Results on Cars and Motorbikes using the \textbf{pairwise model C} described in Section~\ref{sec:cars-motorbikes}.}
\end{figure}

\noindent\textbf{Pairwise model C.} In this experiment, we keep all inliers in both images and randomly add outliers to the second image. The number of outliers varies from $0$ to $40$ by intervals of $5$. We tried the \textbf{pairwise model B} described in Section~\ref{sec:house-hotel} but obtained unsatisfactory matching results (showed in supplementary material). Inspired by the model in~\cite{torresani2013dual}, we propose below a new model that is very simple yet very suited for real-world images. We set the unary terms to $0$ and compute the pairwise terms as
\begin{equation}\label{eq:pairwise-model3}
\cF_{ij}^2 = \eta \delta + (1- \eta) \frac{1- \cos \alpha}{2},
\end{equation} 
where $\eta \in [0,1]$ is a weight constant and $\delta,\alpha$ are computed from $d_1 = \|\overrightarrow{i_1j_1}\|$ and $d_2 = \|\overrightarrow{i_2j_2}\|$ as
\begin{equation}
\delta = \frac{\abs{d_1 - d_2}}{
d_1+d_2},\quad
\cos \alpha = \frac{\overrightarrow{i_1j_1}}{d_1} \cdot \frac{\overrightarrow{i_2j_2}}{d_2}.
\end{equation}
Intuitively, $\cF_{ij}^2$ computes the geometric agreement between $\overrightarrow{i_1j_1}$ and $\overrightarrow{i_2j_2}$, in terms of both length and direction. The above potentials measure the \emph{dissimilarity} between the edges, as thus the corresponding graph matching problem is a \emph{minimization} one. Pairwise potentials based on both length and angle were previously proposed in~\cite{leordeanu2012unsupervised,torresani2013dual} and~\cite{zhou2012factorized}. However, ours are the simplest to compute. In this experiment, $\eta = 0.5$.\\
We match every image pair and report the average in terms of objective value and matching accuracy for each method in Figure~\ref{fig:results-2nd-CarsBikes}. One can observe that ADGM completely outperformed all the other methods.

\begin{figure}[!tb]
    \begin{subfigure}[t]{0.48\linewidth}
    \centering
        \includegraphics[width=\linewidth]{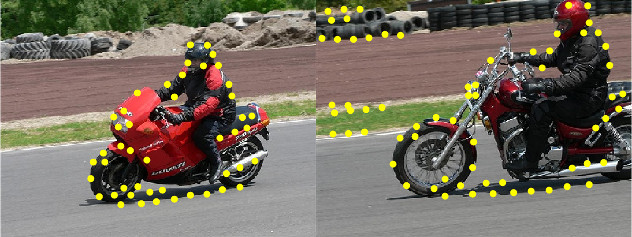}\vspace{-5pt}
        \caption{\small 46 pts vs 66 pts (20 outliers)}
    \end{subfigure}%
    ~
    \begin{subfigure}[t]{0.48\linewidth}
    \centering
        \includegraphics[width=\linewidth]{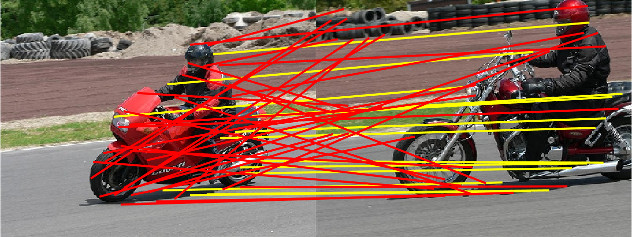}\vspace{-5pt}
        \caption{\small MPM 13/46 (966.2296)}
    \end{subfigure}\\
    ~
    \begin{subfigure}[t]{0.48\linewidth}
    \centering
        \includegraphics[width=\linewidth]{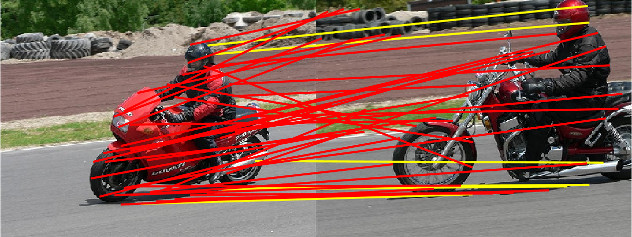}\vspace{-5pt}
        \caption{\small RRWM 6/46 (988.0872)}
    \end{subfigure}%
    ~
    \begin{subfigure}[t]{0.48\linewidth}
    \centering
        \includegraphics[width=\linewidth]{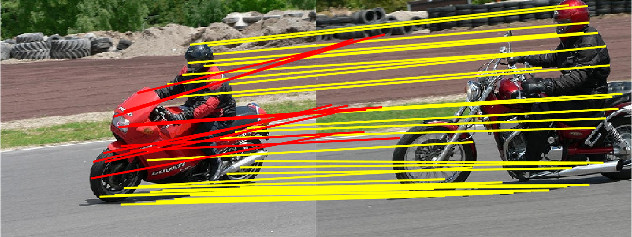}\vspace{-5pt}
        \caption{\small IPFP 35/46 (1038.3965)}
    \end{subfigure}\\
    ~
    \begin{subfigure}[t]{0.48\linewidth}
    \centering
        \includegraphics[width=\linewidth]{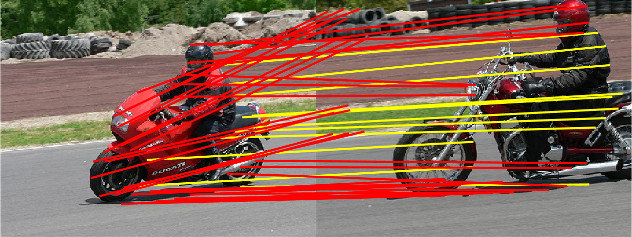}\vspace{-5pt}
        \caption{\small SMAC 11/46 (1028.7961)}
    \end{subfigure}%
    ~
    \begin{subfigure}[t]{0.48\linewidth}
    \centering
        \includegraphics[width=\linewidth]{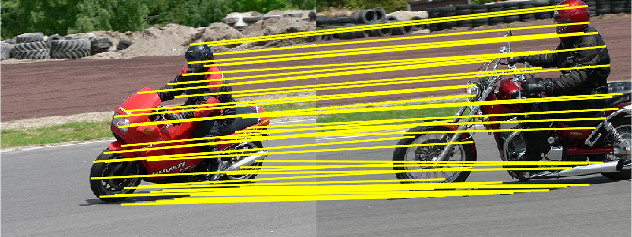}\vspace{-5pt}
        \caption{\small ADGM \textbf{46/46} (\textbf{1043.0687})}
    \end{subfigure}%
    \caption{\label{fig:2nd-demo-CarsBikes}Motorbike matching using the \textbf{pairwise model C} described in Section~\ref{sec:cars-motorbikes}. (Best viewed in color.)}
\end{figure}

\begin{figure}[!tb]
\centering
    \begin{subfigure}[t]{0.48\linewidth}
        \centering
        \includegraphics[width=\linewidth]{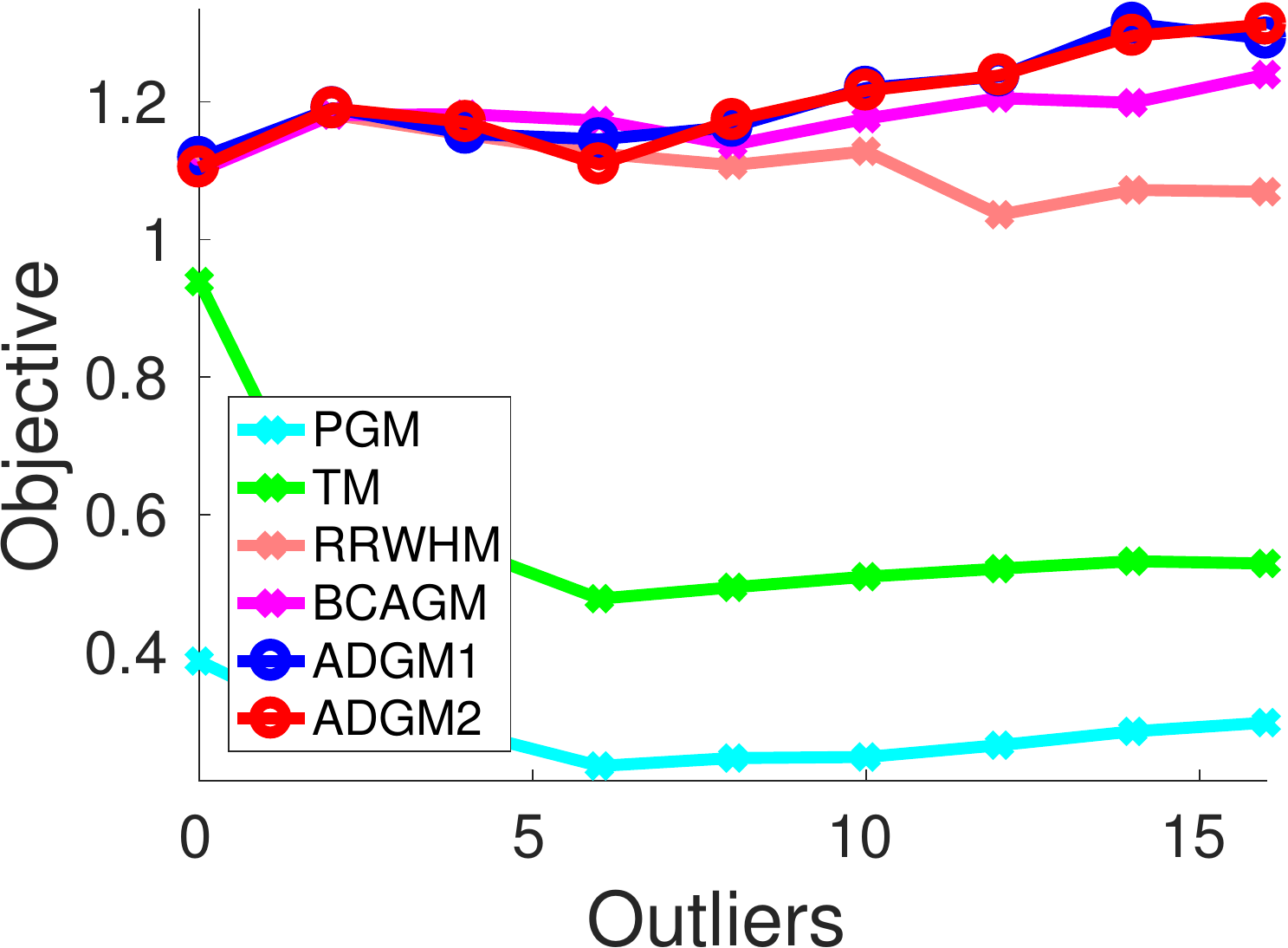}\\
        \includegraphics[width=\linewidth]{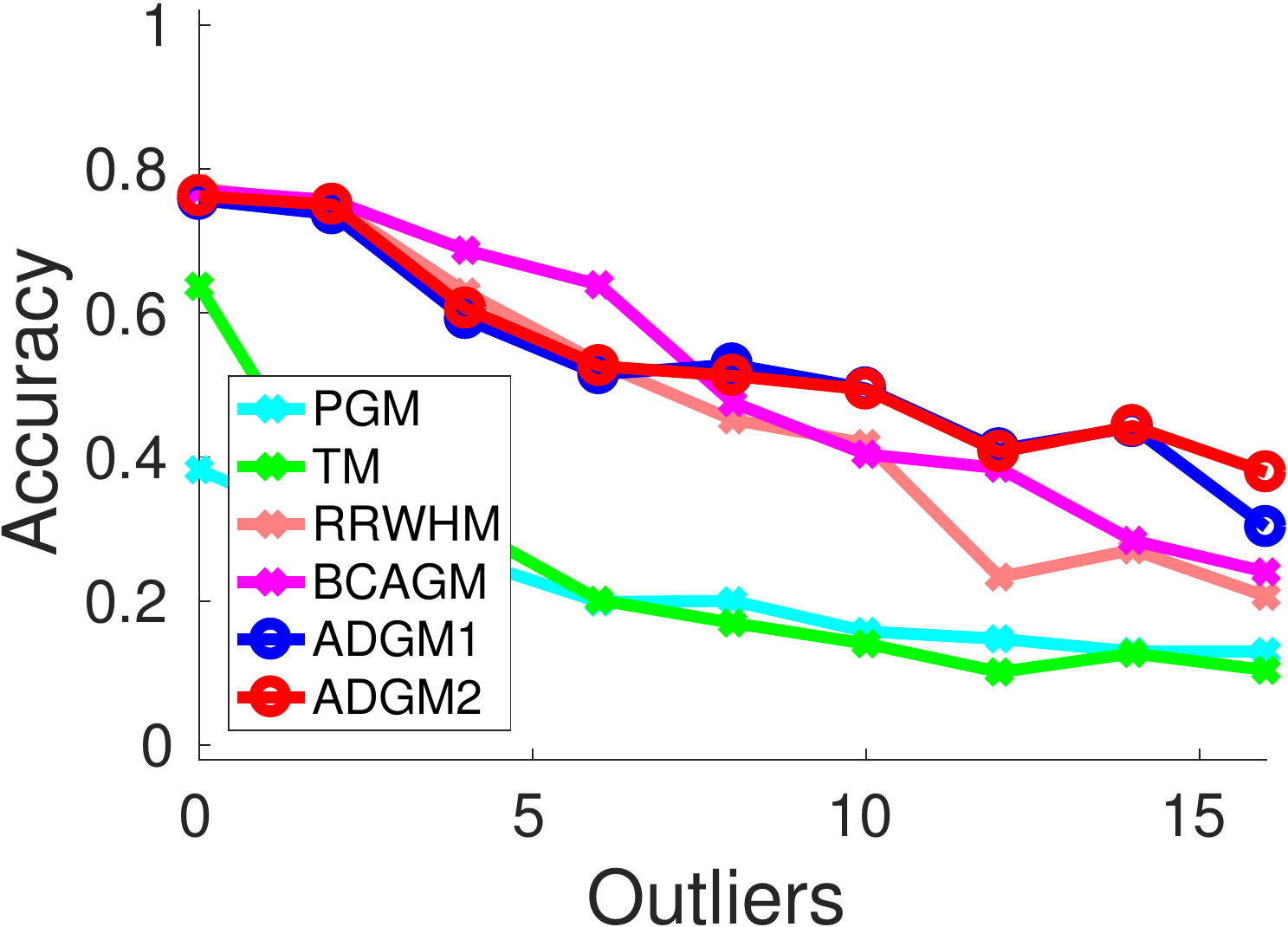}
        \caption{Cars}
    \end{subfigure}%
    ~
    \begin{subfigure}[t]{0.48\linewidth}
        \centering
        \includegraphics[width=\linewidth]{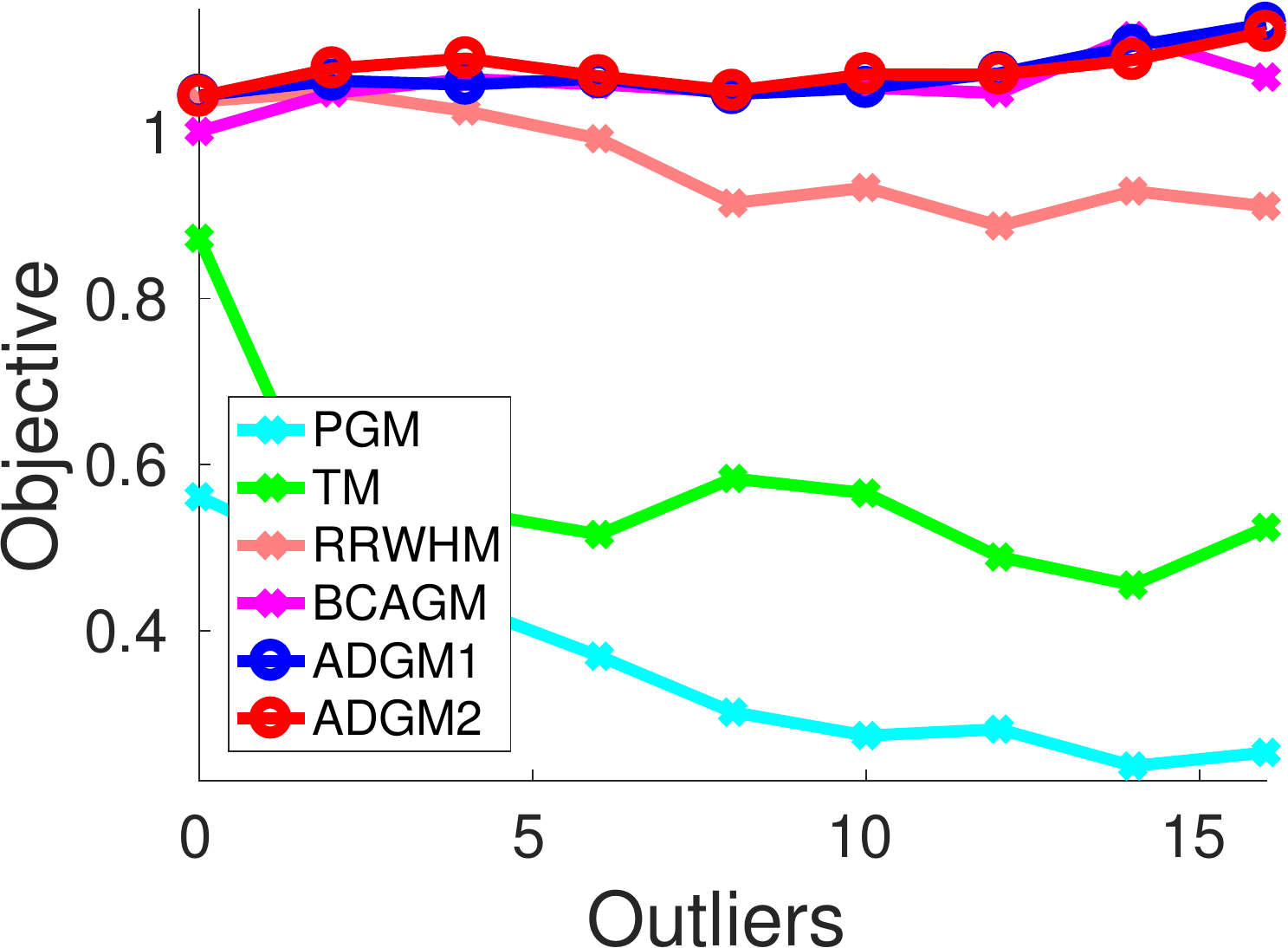}
        \includegraphics[width=\linewidth]{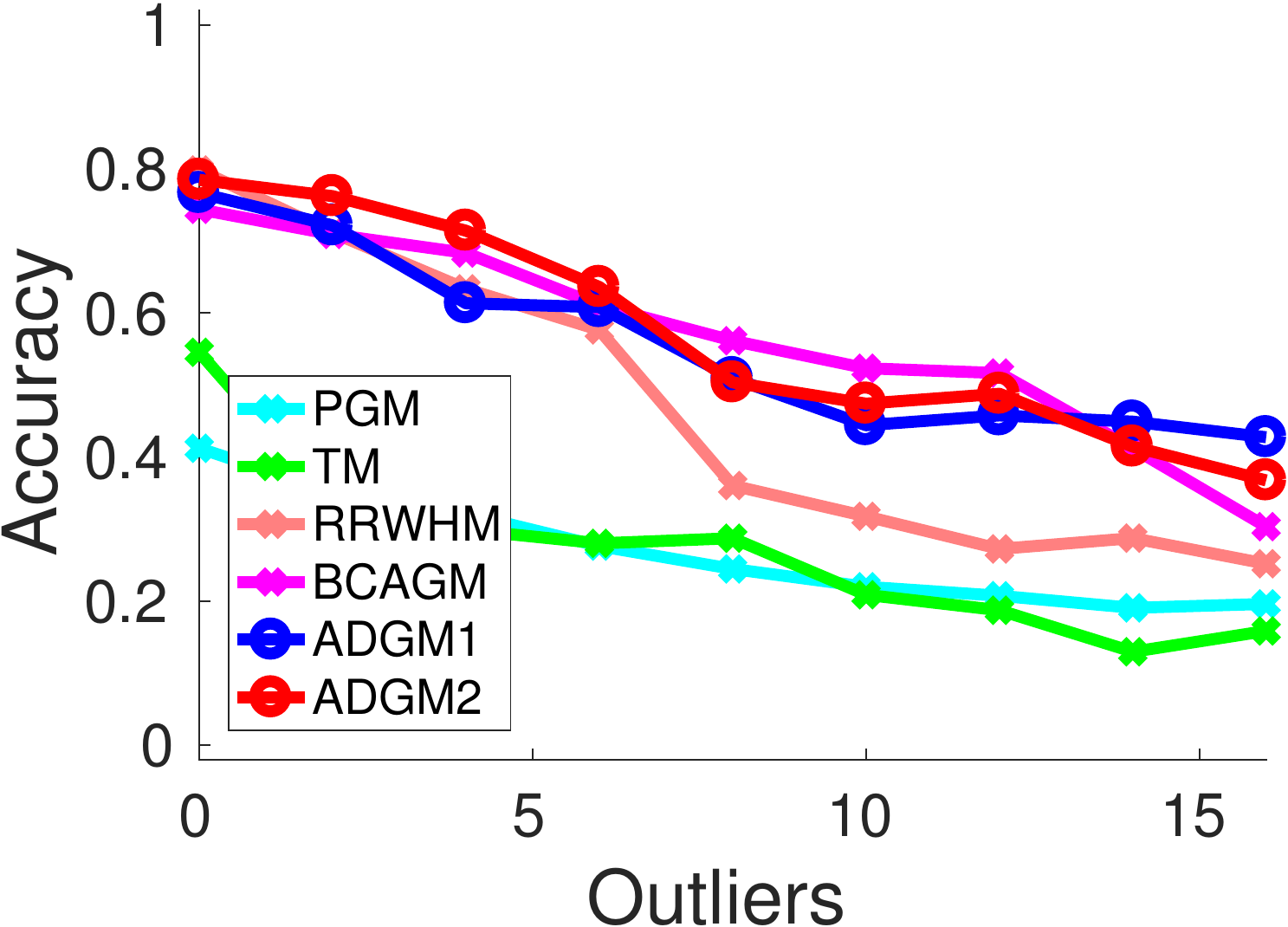}
        \caption{Motorbikes}
    \end{subfigure}%
    \caption{\label{fig:results-3rd-CarsBikes}Results on Cars and Motorbikes using the \textbf{third-order model} described in Section~\ref{sec:cars-motorbikes}.}
\end{figure}

\begin{figure}[!tb]
    \begin{subfigure}[t]{0.48\linewidth}
        \centering
        \includegraphics[width=\linewidth]{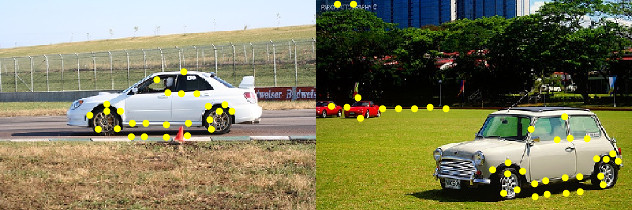}\vspace{-5pt}
        \caption{\small 25 pts vs 36 pts (9 outliers)}
    \end{subfigure}%
    ~
    \begin{subfigure}[t]{0.48\linewidth}
        \centering
        \includegraphics[width=\linewidth]{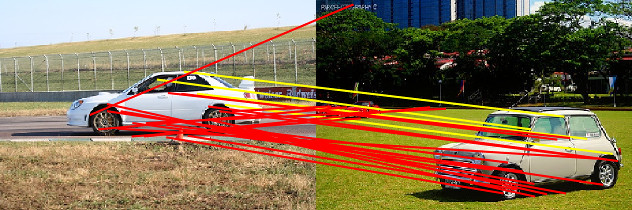}\vspace{-5pt}
        \caption{\small PGM 4/25 (337.8194)}
    \end{subfigure}\\
    ~
    \begin{subfigure}[t]{0.48\linewidth}
        \centering
        \includegraphics[width=\linewidth]{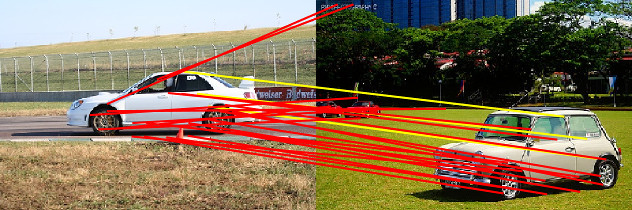}\vspace{-5pt}
        \caption{\small RRWHM 3/25 (1409.832)}
    \end{subfigure}%
    ~
    \begin{subfigure}[t]{0.48\linewidth}
        \centering
        \includegraphics[width=\linewidth]{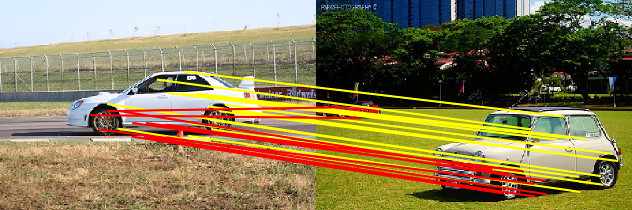}\vspace{-5pt}
        \caption{\small BCAGM 15/25 (1713.487)}
    \end{subfigure}\\
    ~
    \begin{subfigure}[t]{0.48\linewidth}
        \centering
        \includegraphics[width=\linewidth]{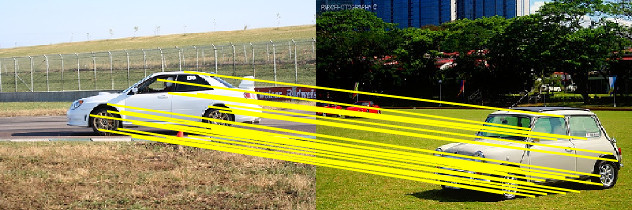}\vspace{-5pt}
        \caption{\small ADGM1 \textbf{25/25} (\textbf{2161.5354})}
    \end{subfigure}%
    ~
    \begin{subfigure}[t]{0.48\linewidth}
        \centering
        \includegraphics[width=\linewidth]{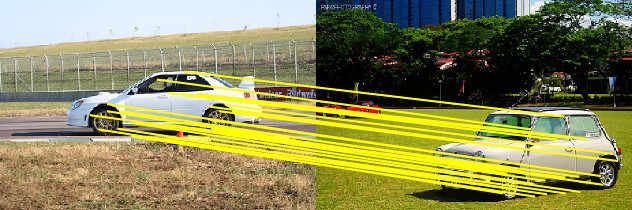}\vspace{-5pt}
        \caption{\small ADGM2 \textbf{25/25} (\textbf{2161.5354})}
    \end{subfigure}%
    \caption{\label{fig:3rd-demo-CarsBikes}Car matching using the \textbf{third-order model} described in Section~\ref{sec:cars-motorbikes}. (Best viewed in color.)}
\end{figure}

\noindent\textbf{Third-order model.} This experiment has the same settings as the previous one, except that it uses a third-order model (the same as in House and Hotel experiment) and the number of outliers varies from $0$ to $16$ (by intervals of $2$). Results are reported in Figure~\ref{fig:results-3rd-CarsBikes} and a matching example is given in Figure~\ref{fig:3rd-demo-CarsBikes}. ADGM did quite well on this dataset. On Cars, both ADGM1 and ADGM2 achieved better objective values than BCAGM in $7/9$ cases. On Motorbikes, ADGM1 beat BCAGM in $5/9$ cases and had equivalent performance in $1/9$ cases; ADGM2 beat BCAGM in $8/9$ cases.

\section{Conclusion and future work}\label{sec:conclusion}
We have presented ADGM, a novel class of algorithms for solving graph matching. Two examples of ADGM were implemented and evaluated. The results demonstrate that they outperform existing pairwise methods and competitive with the state of the art higher-order methods. In future work, we plan to adopt a more principled adaptive scheme to the penalty parameter, and to study the performance of different variants of ADGM. A software implementation of our algorithms are available for download on our website.
\vfill
\noindent\textbf{Acknowledgements.} We thank the anonymous reviewers for their insightful comments and suggestions.

%\newpage

{\small
\bibliographystyle{ieee}
\bibliography{biblio}
}

\end{document}